\documentclass[journal]{IEEEtran}

\IEEEoverridecommandlockouts

\usepackage[table,usenames,dvipsnames]{xcolor}
\usepackage{cite}
\usepackage{lipsum}
\usepackage{tikz}

\usepackage{amsmath} % math
\usepackage{amssymb,amsfonts, amsthm, dsfont, mathtools, bm} % math
\usepackage[linesnumberedhidden, lined, boxed, ruled]{algorithm2e}

\usepackage{float}
\usepackage{graphicx}
\ifCLASSOPTIONcompsoc
\usepackage[caption=false,font=normalsize,labelfont=sf,textfont=sf]{subfig}
\else
\usepackage[caption=false,font=footnotesize]{subfig}
\fi
\usepackage{tabularx, threeparttable}
\usepackage{multirow,multicol}
\def\limsup{\mathop{\lim\sup}\limits}	%
\def\argmax{\mathop{\arg\max}\limits}	%
\DeclareMathOperator{\sinc}{sinc}

\newcommand{\prl}[1]{\mathopen{}\left(#1\right)\mathclose{}}

\newcommand{\crl}[1]{\mathopen{}\left\{#1\right\}\mathclose{}}

\newtheorem{theorem}{Theorem}

\newtheorem{lemma}{Lemma}
\newtheorem*{assumption*}{Assumption}
\newtheorem*{problem*}{Problem}

\title{%\vspace{6mm}
\LARGE \bf
Stochastic Motion Planning under Partial Observability \\ for Mobile Robots with Continuous Range Measurements
}

\author{Ke~Sun,~\IEEEmembership{Student Member,~IEEE,}
        Brent~Schlotfeldt,~\IEEEmembership{Student Member,~IEEE,}\\
        George~Pappas,~\IEEEmembership{Fellow,~IEEE,}
        and~Vijay~Kumar~\IEEEmembership{Fellow,~IEEE,}%
\thanks{We gratefully acknowledge the support of ARL grant ARL DCIST CRA W911NF-17-2-0181.}%
\thanks{The authors are with GRASP Lab, University of Pennsylvania, Philadelphia, PA 19104, USA, {\tt\small\{sunke, brentsc, pappasg, kumar\} @seas.upenn.edu}.}%
}

\begin{document}

\copyright 2020 IEEE. Personal use of this material is permitted.
Permission from IEEE must be obtained for all other uses, in any current or future media, including reprinting/republishing this material for advertising or promotional purposes, creating new collective works, for resale or redistribution to servers or lists, or reuse of any copyrighted component of this work in other works.
\maketitle

\begin{abstract}
  In this paper, we address the problem of stochastic motion planning under partial observability, more specifically, how to navigate a mobile robot equipped with continuous range sensors such as LIDAR.
  In contrast to many existing robotic motion planning methods, we explicitly consider the uncertainty of the robot state by modeling the system as a POMDP.
  Recent work on general purpose POMDP solvers is typically limited to discrete observation spaces, and does not readily apply to the proposed problem due to the continuous measurements from LIDAR.
  In this work, we build upon an existing Monte Carlo Tree Search method, POMCP, and propose a new algorithm POMCP++.
  Our algorithm can handle continuous observation spaces with a novel measurement selection strategy.
  The POMCP++ algorithm overcomes over-optimism in the value estimation of a rollout policy by removing the implicit perfect state assumption at the rollout phase.
  We validate POMCP++ in theory by proving it is a Monte Carlo Tree Search algorithm.
  Through comparisons with other methods that can also be applied to the proposed problem, we show that POMCP++ yields significantly higher success rate and total reward.
\end{abstract}

\section{Introduction}
\label{sec: introduction}

% What is the problem addressed in this paper?
% Why it is important?
% Why it is challenging?
%Motion planning is crucial in enabling a robot to complete a task.
\IEEEPARstart{T}{he} problem of motion planning shows up in almost all robotic applications, including navigating mobile vehicles in complex environments, manipulating robotic arms to grasp objects, controlling micro robots for cellular and chemical delivery, and others.
Depending on the modeling assumptions, motion planning problems can in general be categorized into the following three broad classes:
\begin{itemize}
  \item \textit{Deterministic motion planning} \\
    Problems in this category assume a deterministic motion model, a known initial state, and a known environment, for which an open-loop control policy suffices.
    Such problems are well addressed by search-based algorithms such as A*~\cite{hart1968formal}, or sampling-based algorithms such as RRT, RRT*, and PRM*~\cite{karaman2011sampling}.
  \item \textit{Stochastic model with perfect state information} \\
    In this class, uncertainty of the motion is taken into account, while perfect state information is assumed to be accessible.
    Receding Horizon Control (RHC)~\cite{mayne1990receding}, combined with deterministic motion planning algorithms, directly lends itself to the generation of a feedback control policy for such problems, which implements the idea of certainty equivalent control~\cite[Ch.6.2]{bertsekas2017dynamic}.
    However, it requires one to include motion uncertainty explicitly in cost computation of paths in order to obtain an optimal feedback policy~\cite{melchior2007particle, alterovitz2007stochastic, tedrake2010lqr}.
  \item \textit{Stochastic model with imperfect state information} \\
    Compared with the last two classes, problems in this category provide the most general modeling of a robotic system, which include stochastic motion and measurement models and a prior probability distribution over the state.
    Practitioners typically extend the idea of the separation principle, and rely on a state observer to provide an accurate estimate of the state.
    Then, methods from the second category may be applied.
\end{itemize}

In this paper, we consider problems in the third category, and specifically refer to such problems as \textit{stochastic motion planning}.
Existing practical approaches to solving this problem come with limitations.
Failure to provide accurate state estimate can lead to catastrophic consequences such as collisions.
Furthermore, for estimation methods like visual inertial odometry~\cite{martinelli2012vision}, maintaining observability of the states requires active uncertainty reduction as a part of the planning.
These limitations motivate the exploration of formal approaches to explicitly address the stochastic motion planning problem without resorting to certainty equivalence~\cite{prentice2009belief, van2012motion, silver2010monte, somani2013despot}.

In the context of stochastic motion planning, Partially Observable Markov Decision Processes (POMDPs) \cite{kaelbling1998planning} arise naturally since they provide a mathematical framework which couples stochastic motion and measurement models found in stochastic motion planning problems.
Notwithstanding, the prohibitive computation cost in solving POMDP problems optimally prevents their widespread application.
The technical challenge is twofold.
First, a policy for POMDP problems is known to be a function of belief state instead of simply the state estimate.
Even for a system with finite state space of size $n$, the belief space is continuous and resides in $\mathbb{R}^{n-1}$.
Second, in order to find an optimal policy, one also has to consider all future outcomes, the number of which grows exponentially with the search depth.
Pineau~\cite{pineau2003point} summarizes the two challenges vividly as \textit{curse of dimensionality} and \textit{curse of history}.

% What are the current approaches in general?
% What are the shortcomings of the current approaches?
Many efforts from robotics community have attempted to approximate the stochastic motion planning problem by modeling the systems as Linear systems with Quadratic cost and Gaussian noise (LQG)~\cite{prentice2009belief, van2011lqg}.
Efficient algorithms exist for finding a control policy for such models.
Two major limitations of LQG prevent its wide application to stochastic motion planning problems.
First, LQG methods require a Gaussian state distribution.
In many problems, the belief state of the robot may not be a unimodal Gaussian distribution, especially in ambiguous environments.
Application of LQG methods also requires differentiable objective, dynamical, and measurement functions.
It may not be natural to formulate the task as a differentiable objective function.
More fundamentally, some widely used sensors on robotic platforms, such as LIDAR, lack an explicit and differentiable measurement model.

In the AI community, there is also a strong body of literature on POMDP solvers for general systems~\cite{shani2013survey, ross2008online}, where only a simulator of the system, instead of explicit models, is required.
Recent developments on tree search based general POMDP solvers, Partially Observable Monte Carlo Planning (POMCP)~\cite{silver2010monte} and Determinized Sparse Partially Observable Tree (DESPOT)~\cite{ye2017despot}, have demonstrated effective solutions to large scale problems (PocMan with $10^{56}$ states).

Two difficulties follow immediately in applying the general tree-based POMDP solvers, such as POMCP and DESPOT, to solve stochastic motion planning problems.
First, the general POMDP solvers usually assume a (finite) discrete system, but robotic systems are continuous.
The continuity of the state space should not pose a problem.
Instead of being represented explicitly as a finite dimensional vector, the belief of the state can be approximated with samples.
The continuity of the action space can also be removed by considering only a finite set of motion primitives.
However, it remains unclear how to handle the continuous observation space.
A naive application of the existing tree search methods will result in an overly shallow tree, which is unlikely to produce informed actions that account for a longer horizon, as is necessary in planning problems.

The second difficulty, which is less obvious, comes from estimating the value of the rollout policy.
Consider using deterministic motion planning methods as the rollout policy.
During the rollout phase of tree search based POMDP solvers, one has to find an action sequence based on a sample at a leaf node; apply the actions to the same sample; use the simulated results to estimate the value of the rollout policy.
This is equivalent to the assumption of a known state at the start of rollout phase, resulting in an overly optimistic estimated value, which can lead to incorrect action selection in the tree policy.

% What is our approach in a big picture?
% What are the contributions of this paper?
\textbf{Contributions:}
In this paper, we address the general robotics problem of stochastic motion planning problem with continuous measurements.
We formulate the problem for ground vehicle equipped with a range sensor (LIDAR), whose task is to navigate to a goal on a known occupancy grid map.
We note that the methods proposed here may apply more generally to other robot models and sensor types, but we focus on one problem instance for concreteness.
We solve the navigation problem by building upon POMCP~\cite{silver2010monte}, a tree search method for the solution of general POMDPs, and proposing a new algorithm, POMCP++.
Our major contributions are summarized as follows:
\begin{itemize}
  \item
    Our POMCP++ algorithm addresses the above mentioned difficulties by introducing two major improvements to POMCP.
    First, measurement likelihood of the original system is distorted so that existing measurement branches in the tree can be revisited with nontrivial probability.
    Second, a group of samples, instead of one, is enforced to traverse downstream the same set of nodes in the tree simultaneously.
    The belief at a leaf node can be well represented, resulting in more accurate value estimate of the rollout policy.
    The correct update of the values at the traversed nodes is still guaranteed through the application of importance sampling.
  \item
    Because of the distorted measurement likelihood, the estimated value from the simulation episodes is biased for the current policy.
    Notwithstanding, we prove that such an estimate is unbiased as the number of simulations, and the size of sample group tend to infinity.
    Therefore, the proposed POMCP++ algorithm is shown to be a valid Monte Carlo Tree Search algorithm.
  \item
    We present simulation results that compare POMCP++ with existing approaches that can also be applied to the stochastic navigation problem.
    The results indicate that the POMCP++ algorithm yields a higher summation of discounted reward, the objective to be maximized.
    Furthermore, POMCP++ achieves a higher success rate while reducing the rate of collisions and premature stops outside the goal region.
    In addition to simulations, we demonstrate the performance of POMCP++ on a real robotic platform in an ambiguous hallway environment.
 \end{itemize}

\section{Related Work}
\label{sec: related work}
When addressing stochastic motion problems, most of the works from robotics community assume the robotic system is LQG.
The assumption of the separation principle in LQG allows control and estimation to be performed independently, simplifying the problem.
There is also an independent body of literature from the AI community trying to solve POMDP in its most general form.
In this section, we review literature from these two communities separately.

Earlier works on LQG systems, such as Prentice~\cite{prentice2009belief} and Platt~\cite{platt2010belief}, assume maximum likelihood measurements, which help in identifying a unique posterior distribution, but introduce additional information.
Du Toit~\cite{toit2012robot} proves that maximum likelihood measurement introduces the least amount of information comparing with other future measurement sequences.
van den Berg~\cite{van2011lqg} first figures out the actual dynamics of the belief state of a LQG system.
He shows that the covariance of the posterior Gaussian distribution is fixed for arbitrary future measurements, while the mean follows a Gaussian distribution. i
van den Berg's finding in~\cite{van2011lqg} helps remove the maximum likelihood assumption in later works, such as~\cite{vitus2011closed, van2012motion, agha2014firm, indelman2015planning, sun2016stochastic, rafieisakhaei2017t}.

The LQG approaches can also be categorized based on its origin of the deterministic counterpart.
Censi~\cite{censi2008bayesian} solves the problem with graph search.
Bry~\cite{bry2011rapidly} proposes RRBT by extending RRT*.
More works are based on PRM since the belief roadmap proposed by Prentice~\cite{prentice2009belief}.
A known problem of belief roadmap is the violation of the principle of optimality, due to the lack of strict ordering of covariance matrices.
\cite{censi2008bayesian} and~\cite{bry2011rapidly} address this problem by imposing partial ordering for beliefs.
Agha-Mohammadi~\cite{agha2014firm, agha2015simultaneous} introduces ``belief stablizers'', which assumes the existence of a local controller at each node in PRM that can bring the belief at a node to its neighbor nodes representing other belief.
Shan~\cite{shan2017belief} avoids the necessity of partial ordering by considering the largest eigenvalue of a covariance matrix.
In addition to the above graph search methods, there are algorithms trying to find a control policy through optimization. \cite{van2012motion} and~\cite{sun2016stochastic} directly solve for the feedback control policy through dynamic programming.
\cite{vitus2011closed},~\cite{indelman2015planning}, and~\cite{rafieisakhaei2017t} find the next best action and form a feedback policy through RHC.

Efforts from the AI community try to solve POMDP problems in their most general form without the LQG system assumption.
For finite discrete systems, Pineau~\cite{pineau2003point} first proposes offline point-based method, which forms the basis for many later extensions~\cite{shani2013survey}.
There are two main ideas embedded in the point-based methods which break the \textit{curse of dimensionality} and the \textit{curse of history} respectively.
In the case that an initial belief is known, it is believed that only a small subset of the belief space is reachable.
Therefore, one can only consider the reachable belief space, which can be approximated by a set of belief samples.
The second idea is to exploit the fact that the optimal value function for a finite discrete system is a convex piecewise linear function of the belief, first proved by Sondik~\cite{sondik1971optimal}.
Although the number of piecewise linear regions grows with search depth, one can only keep a fixed number of ``$\alpha$-vectors'' in order to bound the computation complexity.
Extensions of point-based methods, such as~\cite{smith2004heuristic, kurniawati2008sarsop} are usually different in how to choose the belief samples and update the ``$\alpha$-vectors''.

Offline POMDP solvers estimate the value function over the entire belief space.
In contrast, online POMDP solvers focus on estimating the value for the current belief and generating the next best action.
Most of the online POMDP solvers are based on tree search~\cite{ross2008online}, alternating three steps: find the most promising leaf node; expand the node; update the parents of the leaf node through dynamic programming.
Variations of online POMDP solvers~\cite{paquet2005real, ross2007aems} are usually different in what heuristics is applied in finding the next best leaf node to expand.
Developments in sampling based online POMDP solvers, such as POMCP~\cite{silver2010monte} and DESPOT~\cite{somani2013despot}, have demonstrated their effectiveness in solving large scale problems.
It is worth noting the subtle difference between POMCP and tree search methods.
POMCP is based on Monte Carlo Tree Search, where the tree and the pre-defined rollout policy, together, construct a single policy.
As the number of simulations increases, the policy represented by the tree may hopefully converge to the optimal policy.
In contrast, tree search methods, such as DESPOT, embed all policies in the tree.
The job of the algorithm is to identify the optimal policy by accurately estimating the values at the nodes in the tree.

Most of the general POMDP solvers, either offline or online, are limited to finite discrete measurement space.
A few works attempt to relax this limitation.
Hoey~\cite{hoey2005solving} proposes the insight that observations should only be differentiated when they result in different posterior distributions, which provides a metric to partition large, or even continuous, observation space.
Works from~\cite{van2012efficient, chaudhari2013sampling, sunberg2018online} try to solve the continuous POMDPs in general.
van den Berg~\cite{van2012efficient} generalizes the work in~\cite{van2012motion}, focusing on continuous POMDPs with Gaussian noise.
Chaudhari~\cite{chaudhari2013sampling} proposes a formal approach to discretize a continuous POMDP through incremental sampling.
Sunberg~\cite{sunberg2018online} extends POMCP and utilizes progressive widening~\cite{couetoux2011continuous} to handle both continuous action and observation spaces.

A few works try to apply general POMDP solvers for stochastic motion planning problems.
Kurniawati~\cite{kurniawati2011motion} proposes milestone guided sampling, which is, at its root, a point-based method and is still limited to discrete problems.
The specialty of milestone guided sampling is that it employs a PRM-like idea to guide the sampling of belief space.
Lauri~\cite{lauri2016planning} addresses the navigation of a ground vehicle with range sensors using POMCP.
In~\cite{lauri2016planning}, the continuous observation space is discretized by converting range sensor measurements to tuples consisting of cell indices and corresponding occupancy status.
However, the number of possible discrete measurements is still huge, which, again, results in an overly shallow tree.

\section{Problem Formulation}
\label{sec: problem formulation}

\begin{figure}[t]
  \centering
  \subfloat[]{\includegraphics[angle=90, width=0.25\textwidth]{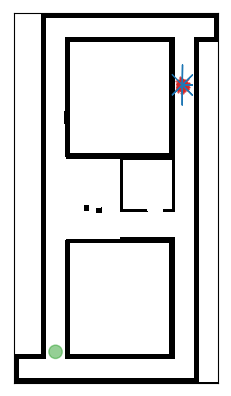}}
  \subfloat[]{\includegraphics[angle=90, width=0.25\textwidth]{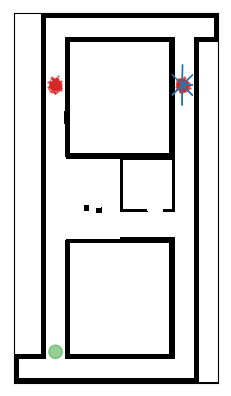}} \\
  \caption{(a) and (b) are representative examples of stochastic motion planning problems, where the task is to navigate a ground vehicle to the target location (green patch) on an occupancy grid map. The state of the robot is unknown. However, the robot is provided with the initial belief (red particles) and has to update its belief through sparse range measurements (blue). (a) shows a configuration with the initial belief consisting of a single mode. The scenario in (b) is more challenging with two modes in the initial belief. The second scenario requires the robot to move out of the corridors in order to collapse the belief into a single mode distribution.}
  \label{fig: stochastic motion planning examples}
\end{figure}

In this work, we address the problem of navigating a ground vehicle to a target location.
Instead of assuming direct access to the perfect state information, the robot has to estimate the state through measurements from onboard range sensors, in this case, a 2-D LIDAR.
We assume the robot is given a known 2-D occupancy grid map, which is necessary for computing measurement likelihood.
\figurename~\ref{fig: stochastic motion planning examples} shows examples of such problem. Formally, the problem is described as follows.

\subsection{Motion Model}
\label{subsec: motion model}
The discrete-time dynamics of the robot is modeled as in~\cite{thrun2005probabilistic}.
\begin{equation}
  \label{eq: stochastic motion model}
  \begin{gathered}
    \begin{pmatrix}
      x_{t+1} \\ y_{t+1} \\ \theta_{t+1}
    \end{pmatrix} =
    \begin{pmatrix}
      x_t \\ y_t \\ \theta_t
    \end{pmatrix} +
    \begin{pmatrix}
      v_t\tau_t \cdot \sinc\left(\frac{\omega_t\tau_t}{2}\right) \cos(\theta_t+\frac{\omega_t\tau_t}{2}) \\
      v_t\tau_t \cdot \sinc\left(\frac{\omega_t\tau_t}{2}\right) \sin(\theta_t+\frac{\omega_t\tau_t}{2}) \\
      \omega_t\tau_t
    \end{pmatrix}, \\
  \end{gathered}
\end{equation}
where $\bm{x} = (x, y, \theta)^\top\in\mathcal{X}$ represents the position and orientation, \textit{i.e.} \textit{state}, of the robot, while $\bm{a} = (v, \omega, \tau)^\top\in\mathcal{A}$ represents the velocity command, \textit{action}.
Furthermore, we note that the \textit{belief} of the robot state is defined as, $\bm{b}\in\mathcal{B}:\mathcal{X} \mapsto \mathbb{R}$, $\bm{b}(\bm{x}) \coloneqq p(\bm{x})$.
To model the uncertainty of system dynamics, the input command $\bm{a}$ is distinguished from the executed command $\tilde{\bm{a}} = \prl{\tilde{v}, \tilde{\omega}, \tilde{\tau}}$,
\begin{equation*}
  \begin{gathered}
    \tilde{v} = v + n_v,\ \tilde{\omega} = \omega + n_\omega + n_\gamma,\ \tilde{\tau} = \tau, \\
    n_v\sim\mathcal{N}(0, \alpha_v(v)^2 + \beta_v(\omega)^2), \\
    n_\omega\sim\mathcal{N}(0, \alpha_\omega(v)^2 + \beta_\omega(\omega)^2), \\
    n_\gamma\sim\mathcal{N}(0, \alpha_\gamma(v)^2 + \beta_\gamma(\omega)^2).
  \end{gathered}
\end{equation*}
where $\alpha_{\{\alpha, \omega, \gamma\}}$ and $\beta_{\{\alpha, \omega, \gamma\}}$ are empirical parameters.
In our work, we restrict the allowable actions to a finite subset $\mathcal{A}_m\subseteq\mathcal{A}$, \textit{i.e.} \emph{motion primitives}, of the action space in order to reduce the complexity of the problem.

\subsection{Measurement Model}
\label{subsec: measurement model}
\textit{Measurements}, $\bm{z}\in\mathcal{Z}$, are ranges of the beams produced by a 2-D LIDAR.
Assuming the range of each beam, $z_i$ $i=0, 1,\dots$ is independent, the measurement likelihood of $\bm{z}$ is the product of the \textit{p.d.f.} of $z_i$'s, \textit{i.e.}
\begin{equation}
  \label{eq: stochastic measurement model}
  p(\bm{z}|\bm{x}, m) = \prod_i p(z_i|\bm{x}, m),
\end{equation}
where $m:\mathcal{X}\mapsto\{0, 1\}$ is a known occupancy grid map.
Therefore, it is sufficient to model a single beam.
In this work, we use the beam model from~\cite{thrun2005probabilistic},
\begin{equation*}
  p(z|\bm{x}, m) =
  \begin{pmatrix}
    z_h \\ z_s \\ z_m \\ z_r
  \end{pmatrix}^\top \cdot
  \begin{pmatrix}
    p_h(z|\bm{x}, m) \\ p_s(z|\bm{x}, m) \\ p_m(z|\bm{x}, m) \\ p_r(z|\bm{x}, m)
  \end{pmatrix}
\end{equation*}
where,
\begin{equation}
  \label{eq: four different measurement noise type}
  \begin{aligned}
    % Normal distribution.
    p_h(z|\bm{x}, m) &=
    \begin{cases}
      \eta_h \cdot \mathcal{N}(z^*, \sigma_h^2), & \text{if } 0\leq z \leq z_{\max}, \\
      0, & \text{otherwise},
    \end{cases} \\
    % Unexpected object.
    p_s(z|\bm{x}, m) &=
    \begin{cases}
      \eta_s \cdot \lambda_s \exp(-\lambda_s z), & \text{if } 0\leq z \leq z^*, \\
      0, & \text{otherwise},
    \end{cases} \\
    % Failures.
    p_m(z|\bm{x}, m) &= \delta(z-z_{\max}), \\
    % Random measurements.
    p_r(z|\bm{x}, m) &=
    \begin{cases}
      \frac{1}{z_{\max}}, & \text{if } 0\leq z \leq z_{\max}, \\
      0, & \text{otherwise},
    \end{cases}
  \end{aligned}
\end{equation}
are four different types of measurement errors due to Gaussian noise, unexpected objects, sensor failure and unexplainable noise.
The probability of each noise type is captured by $(z_h, z_s, z_m, z_r)$ with $z_h+z_s+z_m+z_r=1$.
In Eq.~\eqref{eq: four different measurement noise type}, $z^*$ and $z_{\max}$ are the nominal and maximum range measurement.
For $p_h$ and $p_s$, $\sigma_h$ is the standard deviation for the normal distribution, $\lambda_s$ is the rate of the exponential distribution, and $\eta_{\{h, s\}}$ are the normalization factors.
For $p_m$, $\delta(\cdot)$ is the Dirac delta function.

\subsection{Reward Model}
It is necessary to describe the desired behavior through reward functions.
To encode desired constraints on safety, energy, and total number of actions in our stochastic navigation task, we define a stage reward function as follows:
\begin{equation}
  \label{eq: experiment metric}
  \begin{gathered}
    r(\bm{x}_t, \bm{a}_t, \bm{x}_{t+1}) =
    \begin{cases}
      -5 & \text{if } m(\bm{x}_{t+1}) = 1, \\
      -5 & \text{if } \bm{a}_t=\bm{0} \text{ and } d(\bm{x}_{t+1}, \bm{x}_g)> \epsilon_g, \\
       0 & \text{if } \bm{a}_t=\bm{0} \text{ and } d(\bm{x}_{t+1}, \bm{x}_g) \leq \epsilon_g, \\
      -1 & \text{otherwise},
    \end{cases} \\
    d(\bm{x}_{t+1}, \bm{x}_g) = \sqrt{\|x_{t+1}-x_g\|^2 + \|y_{t+1}-y_g\|^2},
  \end{gathered}
\end{equation}
where $\bm{x}_g$ is the target state.
Cost is induced if the robot keeps moving, which encourages fewer actions.
Colliding with obstacles, or stopping outside the target location, will introduce additional cost.
The accumulation of cost only stops if the robot determines to stop within the target region, which is defined by $\epsilon_g$.
Note the orientation of the $\bm{x}_g$ is ignored in determining task completion.

\begin{problem*}[Stochastic Motion Planning]
	Given the initial belief $\bm{b}_0\in\mathcal{B}$ of a robot, find a control policy $\pi^*:\mathcal{B}\mapsto\mathcal{A}_m$, which maximizes the discounted infinite horizon summation of the stage reward.
  \begin{equation*}
    V_{\pi^*}(\bm{b}_0) =
    \max_{\pi\in\Pi}
    \underset{\substack{\bm{x}_t, \bm{z}_t\\ t=0,1,\dots}}
    {\mathbb{E}}\Big[\sum_{t=0}^\infty \gamma^t r(\bm{x}_t, \bm{a}_t, \bm{x}_{t+1})\Big], \\
  \end{equation*}
  subject to the motion and measurement models given in Eq.~\eqref{eq: stochastic motion model} and Eq.~\eqref{eq: stochastic measurement model}, where $\gamma\in(0, 1)$ is the discount factor.
\end{problem*}

We assume the availability of an estimator able to track the belief of the robot at each time step.
In the following of this work, a particle filer~\cite{fox2002kld} is used to represent the belief using samples of different weights.

\section{Background}
\label{sec: preliminaries}

In this section, we briefly overview POMDP~\cite[Ch.4]{bertsekas2017dynamic} and POMCP~\cite{silver2010monte}.
The stochastic motion planning problem, considered in this work, is concerned with seeking a control policy for a partially observable system, which can be modeled by POMDPs.
POMCP is an extension of Monte Carlo Tree Search (MCTS)~\cite{coulom2006efficient} to solve POMDPs.
The proposed algorithm in this work modifies POMCP addressing stochastic motion planning with continuous measurements.

\subsection{POMDP}
\label{subsec: pomdp}

In POMDP, a system is modeled with an eight-element tuple, $\prl{\mathcal{X}, \mathcal{A}, \mathcal{Z}, t, o, r, b_0, \gamma}$. $\mathcal{X}$, $\mathcal{A}$, $\mathcal{Z}$ are the state, action, and observation spaces.
$t:\mathcal{X} \times \mathcal{A} \times \mathcal{X} \mapsto \mathbb{R}$, $t(\bm{x}, \bm{a}, \bm{x}') \coloneqq p(\bm{x}'|\bm{x}, \bm{a})$, is the transition probability.
$o:\mathcal{X} \times \mathcal{Z} \mapsto \mathbb{R}$, $o(\bm{x}, \bm{z}) \coloneqq p(\bm{z}|\bm{x})$, is the measurement likelihood.
$r:\mathcal{X} \times \mathcal{A} \times \mathcal{X} \mapsto \mathbb{R}$ is a reward function\footnote{The reward function can be also defined as a function of state and action only, \textit{i.e.} $r:\mathcal{X}\times\mathcal{A}\mapsto\mathbb{R}$.}.
In this work, we assume $r(\bm{x}, \bm{a}, \bm{x}')$ is bounded, \textit{i.e.} $|r(\bm{x}, \bm{a}, \bm{x}')|<\infty$.
$\bm{b}_t\in\mathcal{B}:\mathcal{X} \mapsto \mathbb{R}$, $\bm{b}_t(\bm{x}) \coloneqq p(\bm{x}_t)$, represents the distribution of the state at time $t$, while $\mathcal{B}$ is the belief space.
Specially, $\bm{b}_0\in\mathcal{B}_0$ is the initial state distribution. $\gamma\in(0, 1)$ is a discount factor.

The goal in POMDP is to find a control policy $\pi \coloneqq \{\mu_0, \mu_1, \dots, \mu_{N-1}\} \in \Pi$, where $\mu_t:\mathcal{B} \mapsto \mathcal{A}$, such that the discounted summation of reward is maximized\footnote{The posterior distribution of the state is one possible sufficient statistic for the feedback control policy~\cite[Ch.4]{bertsekas2017dynamic}}, \textit{i.e.},
\begin{equation*}
  \begin{gathered}
    V_{\pi}(\bm{b}_0) = \underset{\substack{\bm{x_t}, \bm{z}_t\\t = 0, 1, \dots, N-1}}{\mathbb{E}}
                   \crl{\sum_{t=0}^{N-1} \gamma^t r(\bm{x}_t, \mu_t(\bm{b}_t), \bm{x}_{t+1})}, \\
    \pi^* = \argmax_{\pi\in\Pi} V_{\pi}(\bm{b}_0),\quad V^*(\bm{b}_0) = V_{\pi^*}(\bm{b}_0).
  \end{gathered}
\end{equation*}
In the case that the discounted summation of reward is over infinite horizon, there exists an optimal stationary policy, $\pi^*=\crl{\mu, \mu, \dots}$, which maximizes $\lim_{N\rightarrow\infty} V_{\pi}(\bm{b}_0)$~\cite[Ch.1]{bertsekas2017dynamicvol2}.

\subsection{POMCP}
\label{subsec: pomcp}

\begin{algorithm}[t]

  \DontPrintSemicolon
  \SetCommentSty{emph}
  \SetKwProg{Fn}{Function}{}{end}

  \SetKwFunction{search}{search}
  \SetKwFunction{selectAction}{selectAction}
  \SetKwFunction{simulate}{simulate}

  \SetKwFunction{termination}{termination}
  \SetKwFunction{random}{random}
  \SetKwFunction{process}{process}
  \SetKwFunction{sampleMeasurement}{sampleMeasurement}
  \SetKwFunction{measurementUpdate}{measurementUpdate}
  \SetKwFunction{rollout}{rollout}

  % Function Search.
  \Fn{$\bm{a}$ = \search{$\bm{b}_0$}}{
    $h$ = \textit{NULL}\;
    \While{not \termination{}}{
      \nl $\bm{x}_0$ $\leftarrow$ draw one sample w.r.t. $\bm{b}_0$\;\label{code: sample from the initial belief}
      \nl \simulate{$\bm{x}_0$, $h$, $0$}\;\label{code: simulation following tree policy}
    }
    \KwRet{$\argmax_{\bm{a}} V(h\bm{a})$}
  }

  \BlankLine
  % Function selectAction.
  \nl\Fn{$\bm{a}$ = \selectAction{$h$}}{\label{code: ucb1 action selection}
    \tcp{$c$ is a constant in UCB1 trading-off \\exploitation and exploration.}
    \KwRet{$\argmax_{\bm{a}} V(h\bm{a}) + c\cdot \sqrt{\frac{\log N(h)}{N(h\bm{a})}}$}
  }

  \BlankLine
  % Function simulate.
  \Fn{$r$ = \simulate{$\bm{x}$, $h$, $d$}}{
    \lIf{$\gamma^d < \epsilon$}{\KwRet{$0$}}
    \If{$h$ is NULL}{
      \ForEach{$\bm{a}\in\mathcal{A}$}{
        \nl$h\bm{a}$ $\leftarrow$ $(N_{init}, V_{init})$\;\label{code: tree expansion}
      }
      \nl$r$ = \rollout{$\bm{x}$, $d$}\;\label{code: simulation following rollout policy}
      \KwRet{$r$}
    }

    $\bm{a}$ $\leftarrow$ \selectAction{$h$}\;
    $\bm{x}, r_s$ $\leftarrow$ \process{$\bm{x}, \bm{a}$}\;
    \nl$\bm{z}$ $\leftarrow$ \sampleMeasurement{$\bm{x}$}\;\label{code: sample measurements}

    \nl$r_t$ $\leftarrow$ \simulate{$\bm{x}, h\bm{a}\bm{z} ,d+1$}\;\label{code: simulation following tree policy 2}
    $r \leftarrow r_s + \gamma\cdot r_t$\;
    \nl$N(h) \leftarrow N(h) + 1$\;\label{code: update nodes}
    \nl$N(h\bm{a}) \leftarrow N(h\bm{a}) + 1$\;\label{code: update nodes 2}
    \nl$V(h\bm{a}) \leftarrow V(h\bm{a}) + \frac{1}{N(h\bm{a})}\prl{r-V(h\bm{a})}$\;\label{code: update nodes 3}
  }

  \caption{POMCP}
  \label{alg: pomcp}
\end{algorithm}

Alg.~\ref{alg: pomcp} shows the framework of POMCP.
Some notations in Alg.~\ref{alg: pomcp} are defined as follows.
$h\coloneqq(\bm{a}_0, \bm{z}_0, \dots)$ represents historical information containing a sequence of actions and measurements.
Together with $\bm{b}_0$, $h$ uniquely determines the belief of the state, \textit{i.e.} a node in the tree.
Appending actions and measurements to $h$, such as $h\bm{a}$ or $h\bm{a}\bm{z}$, represents a node in the subtree of $h$.
Specially, $h=\emptyset$ is the root node. $h$'s in the form of $(\bm{a}_0, \bm{z}_0, \dots, \bm{z}_t)$ ended with a measurement represent the belief nodes, while those ended with an action represent the belief-action nodes.
Functions $V(\cdot)$ and $N(\cdot)$ keep track of the estimated value and the number of visits of a node.

For each iteration of the algorithm, a state sample, $\bm{x}_0$ is drawn from the initial belief $\bm{b}_0$ (Alg.~\ref{alg: pomcp} Line~\ref{code: sample from the initial belief}).
The state sample is first simulated recursively following the \textit{tree policy} (Alg.~\ref{alg: pomcp} Line~\ref{code: simulation following tree policy} and~\ref{code: simulation following tree policy 2}).
Once a leaf node in the tree is met, it is expanded (Alg.~\ref{alg: pomcp} Line~\ref{code: tree expansion}).
The sample is forward simulated following the predefined \textit{rollout policy} (Alg.~\ref{alg: pomcp} Line~\ref{code: simulation following rollout policy}) as a continuation.
The traversed nodes are updated based on the simulation results (Alg.~\ref{alg: pomcp} Line~\ref{code: update nodes}-\ref{code: update nodes 2}).
The tree policy is, therefore, implicitly updated due to the addition of the new leaf nodes and the update of the corresponding parent nodes.

\section{POMCP++}
\label{sec: algorithm}

\begin{algorithm*}[ht]
  \begin{multicols}{2}

    \DontPrintSemicolon
    \SetCommentSty{emph}
    \SetKwProg{Fn}{Function}{}{end}

    \SetKwFunction{search}{search}
    \SetKwFunction{simulate}{simulate}
    \SetKwFunction{rollout}{rollout}
    \SetKwFunction{selectAction}{selectAction}
    \SetKwFunction{selectMeasurement}{selectMeasurement}

    \SetKwFunction{random}{random}
    \SetKwFunction{termination}{termination}
    \SetKwFunction{process}{process}
    \SetKwFunction{sampleMeasurement}{sampleMeasurement}
    \SetKwFunction{measurementLikelihood}{measurementLikelihood}
    \SetKwFunction{anytimeAstar}{AnytimeA\textsuperscript{*}}

    % Function Search.
    \Fn{$\bm{a}$ = \search{$\bm{b}_0$}}{
      $h$ = \textit{NULL}\;
      \While{not \termination{}}{
        \nl$\mathcal{S}$ $\leftarrow$ draw $K$ samples w.r.t $\bm{b}_0$\;\label{code: sampling from initial belief}
        \simulate{$\mathcal{S}$, $h$, $0$}\;
      }
      \KwRet{$\argmax_{\bm{a}} V(h\bm{a})$}
    }

    \BlankLine
    % Function rollout
    \nl\Fn{$\bm{r}$ = \rollout{$\mathcal{S}$, $d$}}{\label{code: rollout policy}
      \lIf{$\gamma^d < \epsilon$}{\KwRet{$\bm{0}$}}
      sample $\bm{x} \sim \mathcal{S}$ w.r.t $\bm{w}$\;
      $\{\bm{a}_{r}\}$ $\leftarrow$ \anytimeAstar{$\bm{x}$}\;
      $\bm{r}$ $\leftarrow$ $\bm{0}$\;

      \ForEach{$\bm{x}_i, r_i\in \mathcal{S}, \bm{r}$}{
        \ForEach{$\bm{a}_{r,j} \in \{\bm{a}_r\}$}{
          \lIf{$\gamma^{j+d} < \epsilon$}{break}
          $\bm{x}_i, r$ $\leftarrow$ \process{$\bm{x}_i$, $\bm{a}_{r,j}$}\;
          $r_i$ $\leftarrow$ $r_i + \gamma^j \cdot r$\;
        }

        \If{$\bm{x}_i$ is not at the target location}{
          \tcp{$\bar{r}$ is the average stage reward.}
          $r_i$ $\leftarrow$ $r_i + \gamma^{len(\bm{a}_d)}\cdot \bar{r}$\;
        }
      }

      \KwRet{$\bm{r}$}
    }

    \BlankLine
    % Function selectAction.
    \nl\Fn{$\bm{a}$ = \selectAction{$h$}}{\label{code: epsilon greedy action selection}
      \tcp{\random{} generates random numbers from \\the unit interval with uniform distribution.}
      \eIf{\random{} $>$ $\epsilon_a$}{
        $\bm{a}$ $\leftarrow$ $\argmax_{\bm{a}} V(h\bm{a})$\;
      }{
        uniformly choose $\bm{a}\in\mathcal{A}$\;
      }
      \KwRet{$\bm{a}$}
    }

    \BlankLine
    % Function selectMeasurement.
    \nl\Fn{$\bm{z}$ = \selectMeasurement{$h\bm{a}$, $\mathcal{S}$}}{\label{code: measurement selection}
      \eIf{\random{} $<$ $\big(|C(h\bm{a})|+1\big)^{\epsilon_z}$}{
        sample $\bm{x}\sim \mathcal{S}$ w.r.t $\bm{w}$\;
        $\bm{z}$ $\leftarrow$ \sampleMeasurement{$\bm{x}$}\;
      }{
        uniformly choose $\bm{z}$ s.t. $h\bm{a}\bm{z} \in C(h\bm{a})$\;
      }
      \KwRet{$\bm{z}$}
    }

    \BlankLine
    % Function Simulate
    \Fn{$\bm{r}, \bm{w}$ = \simulate{$\mathcal{S}$, $h$, $d$}}{
      \lIf{$\gamma^d < \epsilon$}{\KwRet{$\bm{0}, \bm{1}$}}
      \If{$h$ is NULL}{
        \ForEach{$\bm{a}\in\mathcal{A}_m$}{
          $h\bm{a}$ $\leftarrow$ $(N_{init}, V_{init})$\;
        }
        $\bm{r}$ = \rollout{$\mathcal{S}$, $d$}\;
        \KwRet{$\bm{r}, \bm{1}$}
      }

      $\bm{a}$ $\leftarrow$ \selectAction{$h$}\;
      $\mathcal{S}, \bm{r}_s$ $\leftarrow$ \process{$\mathcal{S}$, $\bm{a}$}\;
      $\bm{z}$ $\leftarrow$ \selectMeasurement{$h\bm{a}$, $\mathcal{S}$}\;
      $\bm{w}_s$ $\leftarrow$ \measurementLikelihood{$\mathcal{S}$, $\bm{z}$}\;

      $\bm{w}$ $\leftarrow$ $\{w_i\}$ s.t. $(\bm{x}_i, w_i)\in \mathcal{S}$\;
      \ForEach{$w_i, w_{s,i}\in \bm{w}, \bm{w}_s$}{
        $w_i$ $\leftarrow$ $w_i\cdot w_{s,i}$\;
      }
      %\textcolor{blue}{$\bm{w}$ $\leftarrow$ $\bm{w}\cdot \bm{w}_s$}\;

      $\bm{r}_t, \bm{w}_t$ $\leftarrow$ \simulate{$\mathcal{S}$, $h\bm{a}\bm{z}$, $d+1$}\;
      $\bm{r}$ = $\bm{r}_s + \gamma\cdot\bm{r}_t$\;
      %$\bm{w}$ $\leftarrow$ $\bm{0}$\;

      \nl\ForEach{$w_i, w_{s,i}, w_{t,i} \in \bm{w}, \bm{w}_s, \bm{w}_t$}{\label{code: importance sampling}
        $w_i$ $\leftarrow$ $w_{s,i}\cdot w_{t,i}$\;
      }
      %\textcolor{blue}{
      %  \ForEach{$w_i, w_{t,i} \in \bm{w}, \bm{w}_t$}{
      %    $w_i$ $\leftarrow$ $w_i\cdot w_{t,i}$\;
      %  }
      %}
      %\textcolor{blue}{$\bm{w}$ $\leftarrow$ $\bm{w}_s\cdot\bm{w}_t$}\;
      $\bm{w}$ $\leftarrow$ $\bm{w} / \|\bm{w}\|_2$\;
      \nl$r\leftarrow \bm{w}^\top \bm{r}$\;\label{code: importance sampling 2}

      $N(h\bm{a})$ $\leftarrow$ $N(h\bm{a}) + 1$\;
      $V(h\bm{a}) \leftarrow V(h\bm{a}) + \frac{1}{N(h\bm{a})}\prl{r-V(h\bm{a})}$\;

      %\KwRet{$\bm{r}$, $\bm{w}_t$}
      \KwRet{$\bm{r}$, $\bm{w}$}
    }
    \BlankLine

  \end{multicols}

  \caption{POMCP++}
  \label{alg: pomcp++}

\end{algorithm*}

\begin{figure}[th]
  \centering
  \includegraphics[width=\columnwidth]{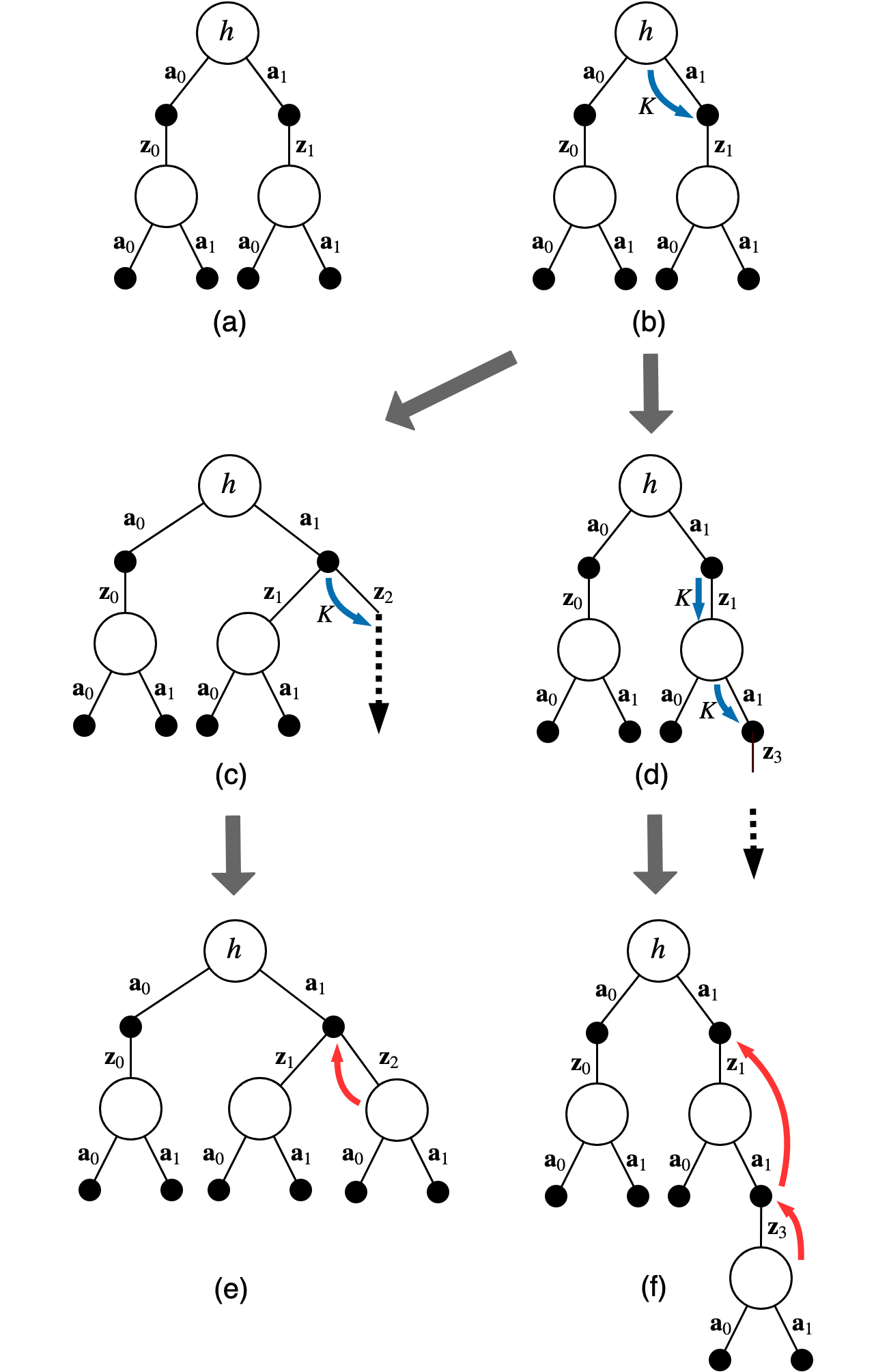}
  \caption{(a) shows a starting tree configuration with two possible actions, $\bm{a}_0$ and $\bm{a}_1$, rooted at $h$. ``$\bigcirc$'' and ``$\bullet$'' denote the belief and belief action nodes respectively. (b) assumes $K$ samples are drawn from the initial belief and $\bm{a}_1$ is selected. The flow of the samples are marked by ``\boldmath$\rightarrow$''. After forward simulating the motion model, different cases happen depending on whether (c) a new measurement branch should be created or (d) an existing measurement branch should be followed. In the case of creating a new measurement branch as in (c), rollout policy is followed afterwards, denoted as ``$\cdots$''. (e) shows the creation of the new leaf nodes. The simulation results are back propagated to the belief action node, shown as ``\boldmath$\rightarrow$''. The difference in (d) and (f) is that action and measurement selection procedure are repeated until a leaf node is encountered. The simulation results are, again, backpropagated to all traversed belief action nodes.}
  \label{fig: tree diagram}
\end{figure}

Alg.~\ref{alg: pomcp++} shows the proposed algorithm POMCP++, with a graphical illustration in \figurename~\ref{fig: tree diagram}.
The overall procedure in POMCP++ follows POMCP in Alg.~\ref{alg: pomcp}.
One obvious modification is the change from UCB1~\cite{auer2002finite} (Alg.~\ref{alg: pomcp} Line~\ref{code: ucb1 action selection}) to $\epsilon$-greedy~\cite[Ch.5]{sutton2018reinforcement} (Alg.~\ref{alg: pomcp++} Line~\ref{code: epsilon greedy action selection}) in action selection, both of which trades off exploitation versus exploration.
Kuleshov~\cite{kuleshov2014algorithms} shows experimentally that the relative performance of the two methods vary with problem setups.
In this work, $\epsilon$-greedy policy is applied in order to ease theoretical analysis.
Other major changes of POMCP++ include
1) replacing forward sampling measurements with measurement selection (Alg.~\ref{alg: pomcp++} Line~\ref{code: measurement selection});
2) using a group of samples (Alg.~\ref{alg: pomcp++} Line~\ref{code: sampling from initial belief}), instead of a single one, to approximate belief while ensuring correct update of nodes through importance sampling;
3) estimating the value of the rollout policy (Alg.~\ref{alg: pomcp++} Line~\ref{code: rollout policy}) using the sample group in order to overcome over-optimism.
Details of the three major changes in POMCP++ are discussed in the following subsections.

In addition to the notations defined for Alg.~\ref{alg: pomcp}, a few more are introduced in Alg.~\ref{alg: pomcp++}.
$\mathcal{S}\coloneqq\crl{(\bm{x}, w)}$ is a set consisting of pairs of states and corresponding weights, approximating the belief.
Function $C(\cdot)$ returns the child nodes of the input node, while $|C(\cdot)|$ represent the number of child nodes.

\subsection{Measurement Selection}
\label{subsec: algorithm measurement selection}
In POMCP, measurements are forward sampled (Alg.~\ref{alg: pomcp} Line~\ref{code: sample measurements}) to determine whether to traverse along the tree or expand a new node.
It is valid for cases of small-size finite discrete measurement space or large ones with concentrated distributions.
In such cases, repeated measurements can be forward sampled with high probability, leading to the growth of the tree in depth.
However, it is no longer true for continuous measurement space.
The probability of generating a repeated measurement is $0$ for a general continuous measurement \textit{p.d.f.}.

To overcome the issue, our proposed algorithm POMCP++ uses a new measurement selection strategy, which distorts the measurement likelihood and enforce the selection of existing measurements.
In the function of measurement selection (Alg.\ref{alg: pomcp++} Line~\ref{code: measurement selection}), a new measurement branch is only created with probability $\prl{|C(h\bm{a})|+1}^{\epsilon_z}$.
Recall that $|C(h\bm{a})|$ is the number of child nodes, \textit{i.e.} existing measurement branches, for the belief-action node $h\bm{a}$.
Otherwise, an existing measurement is enforced.
With $\epsilon_z<0$, a new measurement is less likely to be generated as $C(h\bm{a})$ grows.

A similar notion of measurement selection appears in POMCPOW~\cite{sunberg2018online}.
As action selection, measurement selection in~\cite{sunberg2018online} favors measurement branches that are often visited.
This may result in a biased estimation of the policy value.
In other words, the estimated policy value obtained by following the tree may not be a valid estimation of the actual value obtained by applying the policy to the real system.
The underlying reason is that, unlike actions, the robot has no control over what measurement should be acquired in the future.
Future measurement is completely determined by the measurement likelihood.
In principle, measurement should be forward sampled to ensure that ``measurement selection'' exactly follows the measurement likelihood.
However, this leads to an overly shallow tree, the problem to be addressed in the first place.

In POMCP++, an unbiased policy value estimation is guaranteed by two properties of the proposed measurement selection strategy.
First, for each belief action node, new measurement branches are created \textit{infinitely often (i.o.).} as the number of simulation episodes goes to infinity.
Second, existing measurement branches are uniformly selected in determining the branch to follow further.
These two properties ensure the statistics obtained through existing measurement branches serve as unbiased estimates for the ones obtained with the measurement likelihood.
Appendix~\ref{sec: new measurement is sampled i.o.} provides a proof for the first point, and Appendix~\ref{sec: convergence to optimality} provides a proof for the second.

\subsection{Importance Sampling}
Albeit the enforcement in selecting measurement branches, the selected measurement sequence may not be representative for a single sample from the initial belief.
Instead, $K$ samples are drawn from the initial belief (Alg.~\ref{alg: pomcp++} Line~\ref{code: sampling from initial belief}).
All of the $K$ samples are pushed downstream the tree, traversing the same set of nodes.
Simply averaging the simulation results leads to inaccurate policy evaluation, since each initial state sample has different likelihood of generating the measurement sequence in the traversed path.

The theory of importance sampling comes to the rescue.
Assume the selected action and measurement sequence following the belief node $h_t$ is $\bm{a}_t, \bm{z}_t, \bm{a}_{t+1}, \bm{z}_{t+1}, \dots$.
Then the current policy value at the belief action node $h_t\bm{a}_t$ is,
\begin{equation*}
  \begin{gathered}
    \mathbb{E}_{X_t}\crl{R(X_t, A_t, Z_t)}
    = \int_{X_t} R(X_t, A_t, Z_t)
               p\prl{X_t | A_t, Z_t} d X_t,
  \end{gathered}
\end{equation*}
where $X_t \coloneqq \bm{x}_{t:\infty}$, $A_t \coloneqq \bm{a}_{t:\infty}$, $Z_t \coloneqq \bm{z}_{t:\infty}$ for simplicity.
$R(\cdot)$ computes the discounted summation of reward of a simulation episode.
Applying Bayes rule,
\begin{equation*}
  p\prl{X_t | A_t, Z_t}
  = \frac{p\prl{Z_t|X_t, A_t} p\prl{X_t|A_t}}{p\prl{Z_t|A_t}}.
\end{equation*}
In the spirit of importance sampling, $p\prl{X_t|A_t}$ is the proposal distribution, while $p\prl{X_t|A_t, Z_t}$ is the target distribution.
$p\prl{X_t|A_t}$, the proposal distribution, is approximated by forward sampling.
The measurement likelihood $p\prl{Z_t|X_t, A_t}$ serves as the importance weight. $p\prl{Z_t|A_t}$, as the normalization factor, is constant for all state sequences.
Therefore, the current policy value at node $h_t\bm{a}_t$ can be approximated with,
\begin{equation}
  \label{eq: estimation of belief action nodes}
  \begin{gathered}
      \frac{1}{\eta}\sum_{i=0}^{K-1} p\prl{Z_t|X^i_t, A_t} \cdot
      R(X^i_t, A_t, Z_t),
  \end{gathered}
\end{equation}
with $\eta=\sum_i p\prl{Z_t|X^i_t, A_t}$ (Alg.~\ref{alg: pomcp++} Line~\ref{code: importance sampling}-~\ref{code: importance sampling 2}).

\subsection{Rollout Policy}
It is desirable to take advantage of the well-developed deterministic motion planning methods to design the rollout policy.
However, since such deterministic motion planning methods are usually state-dependent, direct application of them in the framework of POMCP or POMCPOW leads to over-optimism.
More specifically, in POMCP and POMCPOW, a single state sample, $\bm{x}_0$, is simulated from the root to a leaf node to obtain $\bm{x}_t$, which is then used to find a deterministic action sequence, $\bm{a}_{t:T}^r$.
$T$ is implicitly determined by the length of the deterministic action sequence.
In the case that a feasible rollout policy cannot be found, $T$ can be set large enough so that $\gamma^T \ll 1$ and the reward-to-go beyond $T$ can be ignored.
The reward-to-go of the rollout policy is estimated with $R(\bm{x}_{t:T}, \bm{a}_{t:T}^r)$, whose actual value is,
\begin{equation*}
  \begin{gathered}
    \underset{p\prl{\bm{x}_{t:T}|\bm{a}_{t:T}^r}}{\mathbb{E}} \crl{R\prl{\bm{x}_{t:T}, \bm{a}_{t:T}^r}} \\
    = \int_{\bm{x}_{t:T}} R\prl{\bm{x}_{t:T}, \bm{a}_{t:T}^r} p\prl{\bm{x}_{t:T}|\bm{a}_{t:T}^r} d\bm{x}_{t:T}.
  \end{gathered}
\end{equation*}
Note that $\bm{x}_t$ may be from belief with high uncertainty at the leaf node.
Since $\bm{a}_{t:T}^r$ is found using $\bm{x}_t$ in the first place, the rollout action sequence is expected to obtain high reward-to-go starting from $\bm{x}_t$.
However, $\bm{a}_{t:T}^r$ may not perform well with states that are significantly different from $\bm{x}_t$.
Therefore, $\underset{p\prl{\bm{x}_{t:T}|\bm{a}_{t:T}^r}}{\mathbb{E}} \crl{R\prl{\bm{x}_{t:T}, \bm{a}_{t:T}^r}}$ may be much less than $R(\bm{x}_{t:T}, \bm{a}_{t:T}^r)$.
If $R(\bm{x}_{t:T}, \bm{a}_{t:T}^r)$ is then back propagated to the parent nodes, the high reward-to-go may mislead action selection in later steps to favor this branch albeit the high uncertainty at the leaf node.
In other words, the implicit assumption of perfect state information of $\bm{x}_t$ should be accused for the over-optimism.

Because of the measurement selection and importance sampling introduced previously, one has $K$ state samples when reaching a leaf node.
One of the $K$ samples, $\bm{x}_t$, is chosen and used to find the open-loop action sequence $\bm{a}_{t:T}^r$.
The estimated reward for the rollout policy is then $\sum_{i=0}^{K-1} R\prl{\bm{x}_{t:T}^i, \bm{a}_{t:T}^r}$ (Alg.~\ref{alg: pomcp++} Line~\ref{code: rollout policy}).
In this work, we use $A^*$, with inflation factor of the heuristic set to infinity~\cite{likhachev2004ara}, to find the rollout policy.
Although the action sequence could be sub-optimal, it ensures maximum efficiency of the rollout policy construction.

Removal of the implicit perfect state information assumption at the leaf nodes helps address the underlying active localization problem.
Active localization requires the robot to actively reduce the state uncertainty in order to complete the task with high confidence.
Since the action sequence $\bm{a}_{t:T}^r$ is constructed with $\bm{x}_t$, in POMCP++, the rollout policy only attains high reward-to-go when all samples are close to $\bm{x}_t$, which requires the belief at leaf nodes to have low uncertainty.

\section{Experiments}
\label{sec: experiments}

In the experiment section, we compare the performance of POMCP++ with RHC, DESPOT~\cite{ye2017despot}, and POMCPOW~\cite{sunberg2018online} through simulations.
The idea of certainty equivalence control and separation principle is often extended and applied in the receding horizon control framework to solve the stochastic motion planning problem in practice.
In the following, methods of this kind is called RHC for short.
Particularly, to solve the proposed problem, one state sample is drawn from the belief at each step, which is regarded as the true state.
$A^*$, with inflated heuristics~\cite{likhachev2004ara}, is then used to find the next action assuming a deterministic motion model.
DESPOT~\cite{somani2013despot} is the state-of-the-art online POMDP solver designed for discrete systems.
The simulation results of DESPOT serve as the benchmark for naively applying discrete POMDP solvers to continuous systems.
POMCPOW~\cite{sunberg2018online}, also an extension of POMCP, is the most similar to our work.
Comparison with POMCPOW shows the importance of the improvements in POMCP++.

There are minor modifications to POMCPOW and DESPOT to adapt to the proposed problem.
POMCPOW is designed to handle continuous action space in addition to continuous state and observation spaces.
The action space in the problem setup of this work is discrete and finite, consisting of motion primitives.
Therefore, we apply the same $\epsilon$-greedy action selection strategy (Alg~\ref{alg: pomcp++} Line~\ref{code: epsilon greedy action selection}) in the POMCPOW framework.
For both DESPOT and POMCPOW, the same rollout policy as in Alg.~\ref{alg: pomcp++} is used.

Sec.~\ref{subsec: tests in simulated environments} evaluates different methods in relatively simple simulated environments, where the optimal policy can be anticipated.
The performance of the methods are compared in terms of the summation of discounted reward.
The discounted reward is the objective to be maximized, and is widely used as a metric to quantify the performance of general POMDP solvers in the AI literature.
In contrast, Sec.~\ref{subsec: tests in hallway environments} evaluates the methods in a more realistic hallway environment.
The performance of the methods are compared in terms of metrics that are of more interest to robotics researchers, such as the rate of collisions, total number of actions, the entropy of the final distribution, \textit{etc}.
In addition, we compare the performance of POMCPOW and POMCP++ under different per-step planning time constraints.
Finally, we demonstrate the usage of POMCP++ by transitioning the simulation in the hallway environment to hardware experiments.

\subsection{Tests in simulated environments}
\label{subsec: tests in simulated environments}

In the simulated environments, the cell size of the used occupancy grid maps is set to $0.1m$.
The action space consists of six different motion primitives, $\mathcal{A}_m = \{0, 0.2\}m/s \times \{-\frac{\pi}{2}\, 0, \frac{\pi}{2}\}rad/s \times \{0.5\}s$.
The robot models a differential drive ground vehicle with forward motion only.
The range sensor installed on the robot has seven beams covering $2\pi$ uniformly.
Each beam has a maximum range of $z_{\max} = 0.3m$.
$\epsilon_g$ in Eq.~\eqref{eq: experiment metric}, the radius of the goal region, is $0.05m$.\footnote{The setup is merely for the ease to convert quantities to metric length units. All related quantities can be scaled proportionally.}
$\gamma$, the discount factor, is set to $0.99$.

%The hyper-parameters for each method are chosen from cross validation.
Table~\ref{tab: hyperparameters for different methods} shows the range and the selected values for parameters of each method.
See~\cite{somani2013despot} and~\cite{sunberg2018online} for the functionality of the parameters.
It should be specially noted that, for DESPOT, \textit{tree depth} is set to be well beyond the maximum number of steps to reach the goal.
\textit{per-step planning time} is also chosen to be large enough such that time limitation is seldom used as a termination criterion for each planning step.
\begin{table}[t]
  \centering
  \begin{threeparttable}[t]
    \caption{Hyper-parameters for different methods}
    \label{tab: hyperparameters for different methods}
    \renewcommand{\arraystretch}{1.5}
    \begin{tabular}{c|l}
      \hline\hline
      \multirow{3}{5em}{DESPOT}
      & $\text{tree depth}=\{10, \bm{50}, 100\}$ \\
      & $\text{scenario \#}=\{30, 100, \bm{500}, 1000\}$ \\
      & $\text{per-step planning time}=\{\bm{30}\}$ \\
      \hline
      \multirow{4}{5em}{POMCPOW}
      & $\epsilon_a = \{0.05, 0.1, 0.2, \bm{0.3}\}$ \\
      & $\alpha=\{0.1, 0.5, \bm{1}, 2\}$ \\
      & $\kappa=\{\bm{0.1}, 0.5, 1, 5\}$ \\
      & $N=\{1000, \bm{3000}, 5000\}$ \\
      \hline
      \multirow{4}{5em}{POMCP++}
      & $\epsilon_a=\{0.05, \bm{0.1}, 0.2, 0.3\}$ \\
      & $\epsilon_z=\{\bm{-1}, -3, -5, -7\}$ \\
      & $K=\{16, 32, \bm{64}\}$ \\
      & $N=\{\bm{3000}\}$ \\
      \hline\hline
    \end{tabular}
  \end{threeparttable}
\end{table}

\subsubsection{Fixed maps}
\label{subsubsec: simulations on fixed maps}
\begin{figure*}
  \centering
  \subfloat[]{\includegraphics[width=0.2\textwidth]{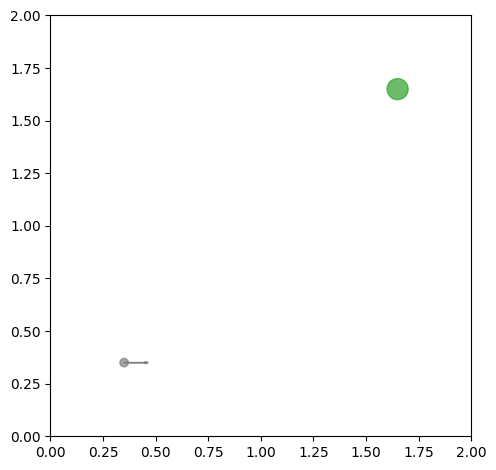}\label{subfig: the special map2}}
  \subfloat[]{\includegraphics[width=0.2\textwidth]{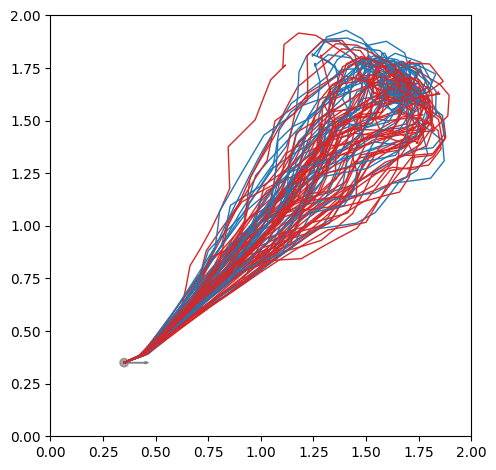}\label{subfig: path of dmpp on the special map2}}
  \subfloat[]{\includegraphics[width=0.2\textwidth]{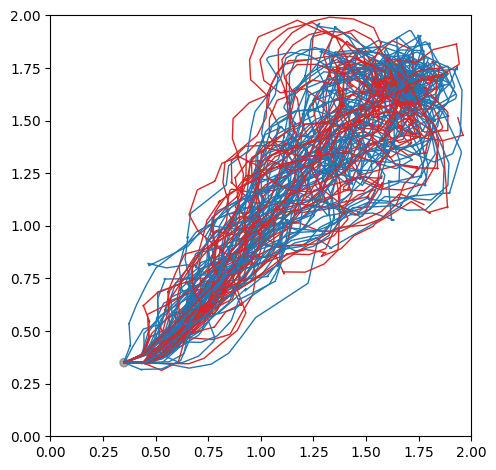}\label{subfig: path of despot on the special map2}}
  \subfloat[]{\includegraphics[width=0.2\textwidth]{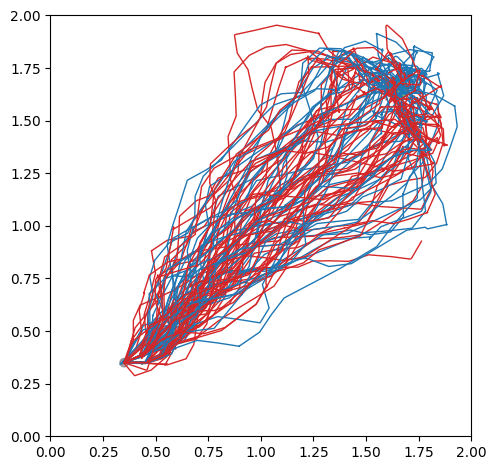}\label{subfig: path of pomcpow on the special map2}}
  \subfloat[]{\includegraphics[width=0.2\textwidth]{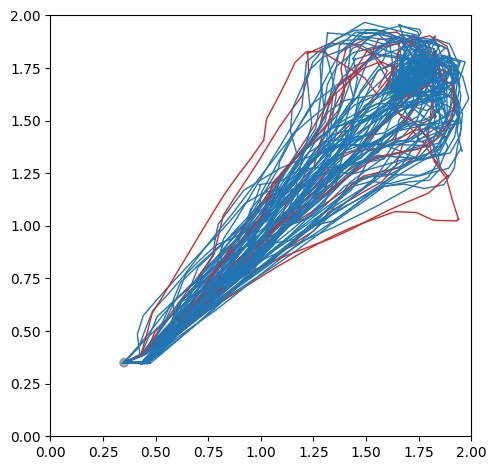}\label{subfig: path of pomcp on the special map2}} \\
  \subfloat[]{\includegraphics[width=0.2\textwidth]{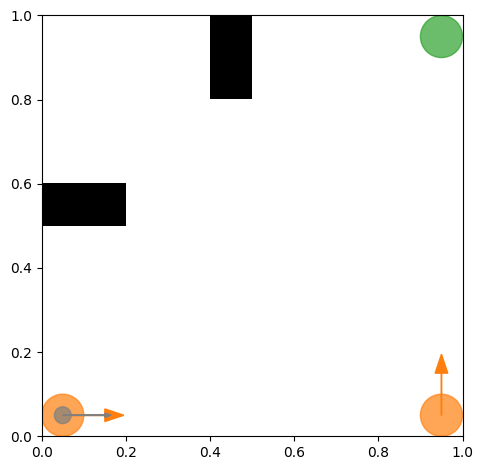}\label{subfig: the special map}}
  \subfloat[]{\includegraphics[width=0.2\textwidth]{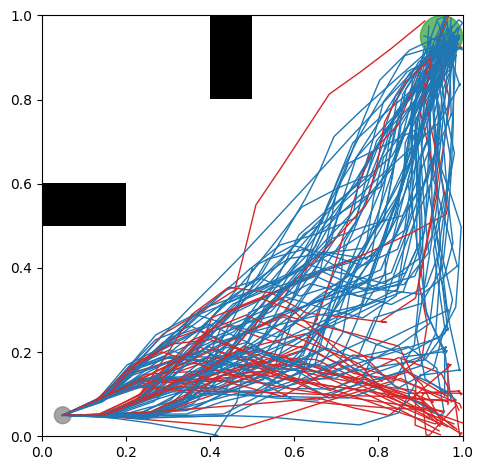}\label{subfig: path of dmpp on the special map}}
  \subfloat[]{\includegraphics[width=0.2\textwidth]{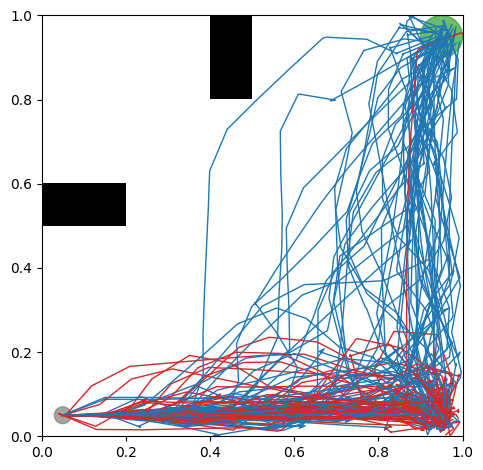}\label{subfig: path of despot on the special map}}
  \subfloat[]{\includegraphics[width=0.2\textwidth]{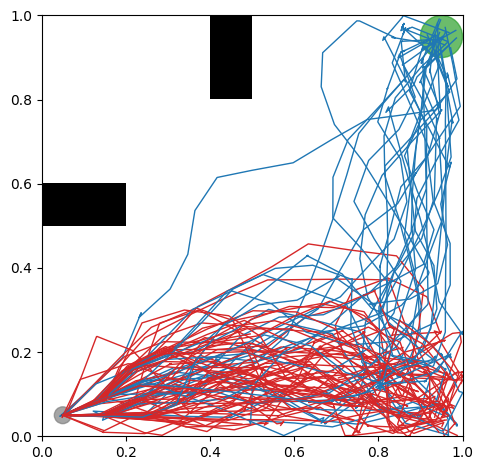}\label{subfig: path of pomcpow on the special map}}
  \subfloat[]{\includegraphics[width=0.2\textwidth]{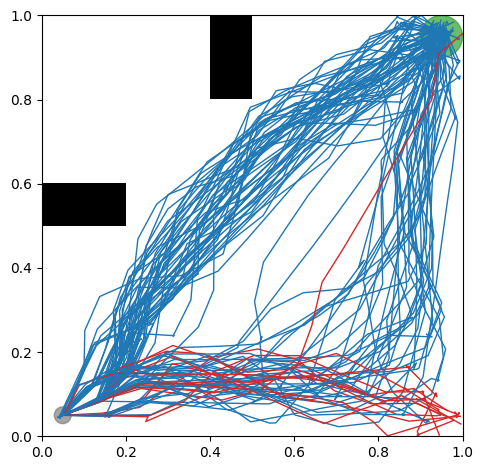}\label{subfig: path of pomcp on the special map}} \\
  \subfloat[]{\includegraphics[width=0.5\textwidth]{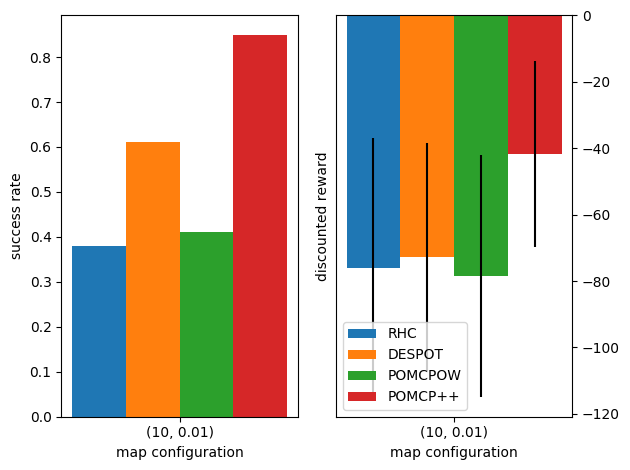}\label{subfig: the special map2 stats}}
  \subfloat[]{\includegraphics[width=0.5\textwidth]{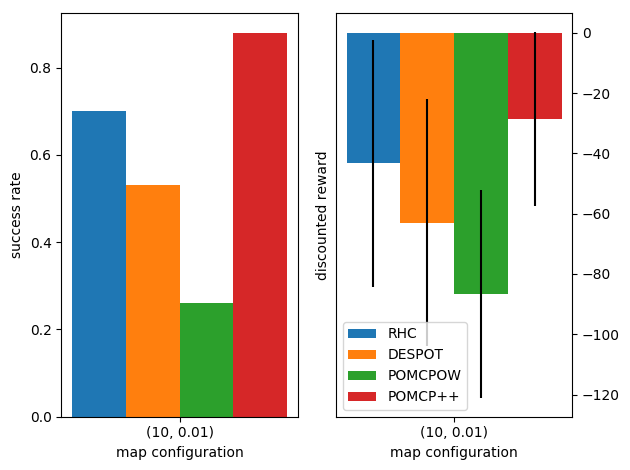}\label{subfig: the special map stats}} \\
  \caption{(a) and (f) show the initial configurations for the experiments in Sec.~\ref{subsubsec: simulations on fixed maps}. Occupied cells in the maps are marked with black squares. The Green circular region is the target location. The gray circle is the initial state of the robot. The orange circles with the arrows in (f) are the two modes in the initial belief. (b), (c), (d), and (e) are the paths executed with RHC, DESPOT, POMCPOW, and POMCP++ in setup (a) over the $100$ experiments. Blue lines are the paths for the successful trials, while red lines are the failed ones. The paths executed by the methods in setup (f) are shown in (g), (h), (i), and (j). (k) and (l) are the success rate and average discounted reward with standard deviation obtained through the $100$ experiments for setup (a) and (f) respectively.}
  \label{fig: experiments on the special maps}
\end{figure*}
In this section, we show both qualitative and quantitative results of applying the methods on two fixed representative maps.
In the first setup, \figurename~\ref{subfig: the special map2}, we use a $20\times 20$ map with no obstacles.
The initial state is known to the robot.
In order to reach the target location with high confidence, it is expected that the robot will take advantage of the map boundaries for localization.
The second setup, \figurename~\ref{subfig: the special map}, is a $10\times 10$ map, designed to be more challenging in that the initial distribution is ambiguous, consisting of two modes located at the bottom corners.
The obstacles in this setup are also set in a way such that the robot is not able to remove the state ambiguity passively, but has to actively approach the obstacles.
We run each method $100$ times on the two setups to obtain the average performance.

\figurename~\ref{subfig: the special map2 stats} and~\ref{subfig: the special map stats} show the success rate and the average discounted reward of different methods.
More insightful observations can be obtained by plotting the paths, \figurename~\ref{fig: experiments on the special maps}.
In the first setup, the proposed POMCP++ is able to reliably utilize the right upper corner to reduce the uncertainty accumulated during the movement, and then come back to the target location.
This capability explains the high success rate of POMCP++.
For the second setup, an immediate observation is that all methods try to go along the bottom and right side of the map.
This is due to the best-first-search nature of the algorithms.
Going along the bottom edge is the fastest way to move one of the belief modes to the target location.
Occasionally, the planning algorithms may wrongly choose to stop at this moment.
More often, exploring the policies helps the algorithms realize that moving along the right side can further reduce the uncertainty and eventually leave a single belief mode stopping at the target.
A more interesting observation is that the proposed POMCP++ is able to find another cluster of paths by moving the robot diagonally across the map.
The paths in this cluster are able to remove the ambiguity in the belief within the first few steps by utilizing the obstacles on the left.
Although such paths deviate from the ``promising'' path (going right, then up) of the short horizon, they obtain higher overall reward.

\subsubsection{Random maps}
\label{subsubsec: simulations on random maps}
\begin{figure*}
  \centering
  \subfloat[]{\includegraphics[width=0.5\textwidth]{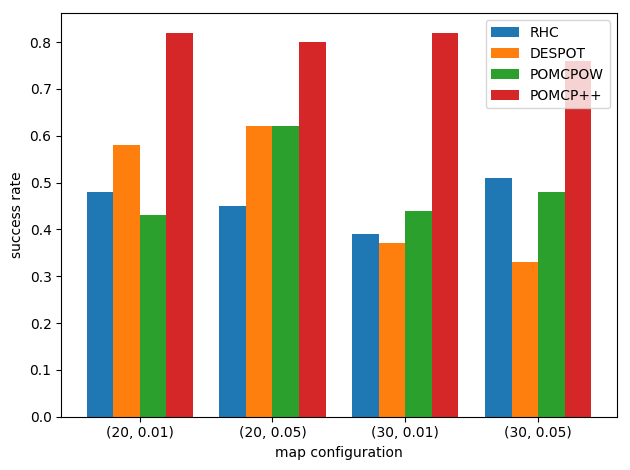}\label{subfig: success rate on random maps}}
  \subfloat[]{\includegraphics[width=0.5\textwidth]{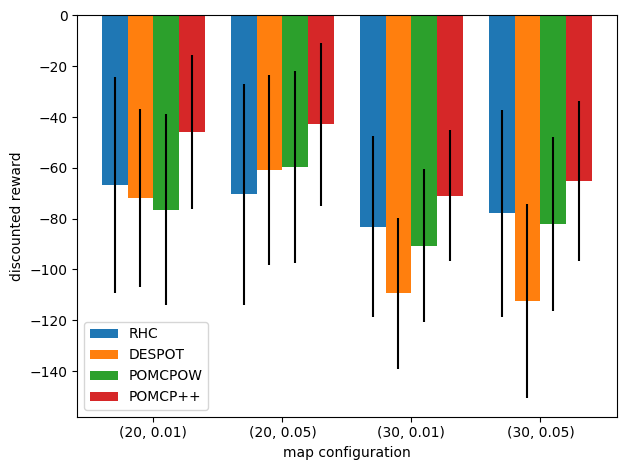}\label{subfig: reward for all trails on random maps}}
  \caption{For each method and map configuration, the performance of the method is measured with $100$ experiments on randomly generated maps. (a) The success rate of different methods. Recall Eq.~\eqref{eq: experiment metric}, a trial is successful when the robot actively chooses to stop around the target location. (b) The average summation of discounted reward with standard deviation. For failure trials, it is assumed that the robot will take feasible actions (with reward $-1$ per step) onwards but never reach the target location.}
  \label{fig: experiments on random maps}
\end{figure*}
In order to evaluate the average performance of different methods, they are also compared using random maps with different configurations.
Four map configurations are included by permuting different map size, $\crl{20\times20, 30\times30}$, and obstacle density, $\crl{0.01, 0.05}$.
$100$ experiments are performed for each method and map configuration.
In each experiment, a random map is generated as per the map configuration.
For all experiments, the robot starts with a known initial state.

\figurename~\ref{fig: experiments on random maps} shows the success rate and summation of discounted reward for different methods.
POMCP++ is able to achieve much higher success rate and reward, constantly outperforming the other three methods.
It should be also noted that RHC performs reasonably well in this experiment, better than the POMDP-based methods, DESPOT and POMCPOW, on some map configurations.

\subsection{Tests in hallway environments}
\label{subsec: tests in hallway environments}

\begin{table*}[t]
  \centering
  \begin{threeparttable}[t]
    \caption{Comparison of RHC, POMCPOW, and POMCP++ in a hallway environment}
    \label{tab: comparison of pomcp++ and rhc}
    \renewcommand{\arraystretch}{1.5}
    \begin{tabular}{c|c|c|c|c|c|c}
      \hline\hline
      initial belief
      & method
      & success/collision/wrong stop/timeout\tnote{1} [\%]
      & total actions\tnote{2}
      & total distance\tnote{2} [$m$]
      & distance to target\tnote{2} [$m$]
      & final entropy\tnote{2} \\
      \hline
      % single mode
      \multirow{3}{5em}{single mode}
      & RHC     & $    0.66 /    0.32 /    0.02 /\bm{0.00}$ & $\bm{66.48}$ & $\bm{29.13}$ & $    0.45 $ & $    2.92 $\\
      & POMCPOW & $    0.78 /    0.08 /    0.10 /    0.04 $ & $    74.36 $ & $    29.64 $ & $    0.41 $ & $    2.56 $\\
      & POMCP++ & $\bm{0.84}/\bm{0.00}/\bm{0.00}/    0.16 $ & $    75.69 $ & $    29.51 $ & $\bm{0.31}$ & $\bm{2.53}$\\
      \hline
      % two modes
      \multirow{3}{5em}{two modes}
      & RHC     & $    0.58 /    0.38 /     0.04 /\bm{0.00}$ & $\bm{66.62}$ & $\bm{29.07}$ & $    0.44 $ & $    2.94 $\\
      & POMCPOW & $    0.34 /    0.36 /     0.30 /\bm{0.00}$ & $    74.18 $ & $    29.66 $ & $    0.39 $ & $\bm{2.40}$\\
      & POMCP++ & $\bm{0.90}/\bm{0.00}/ \bm{0.02}/    0.08 $ & $    73.38 $ & $    29.28 $ & $\bm{0.26}$ & $    2.46$\\
      \hline\hline
    \end{tabular}
    \begin{tablenotes}
      \item [1] There are three causes of a failure trial, collision, wrong stop (stops more than $0.5m$ away from the target), or timeout (uses over $100$ steps).
      \item [2] These data is averaged over successful trials only.
    \end{tablenotes}
  \end{threeparttable}
\end{table*}

\begin{figure*}[t]
  \centering
  \subfloat[]{\includegraphics[angle=90, width=0.33\textwidth]{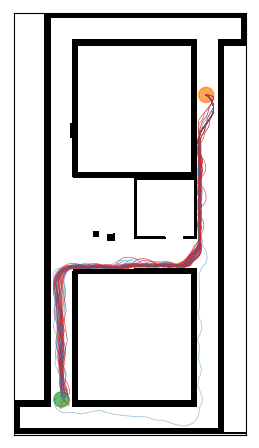}}
  \subfloat[]{\includegraphics[angle=90, width=0.33\textwidth]{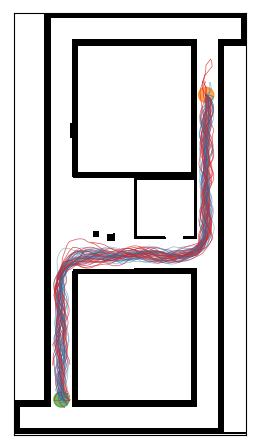}}
  \subfloat[]{\includegraphics[angle=90, width=0.33\textwidth]{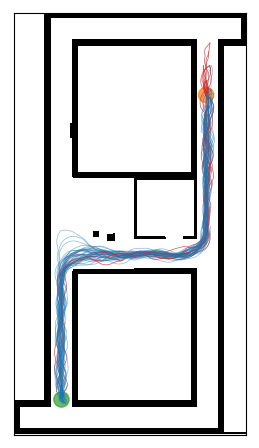}} \\
  \subfloat[]{\includegraphics[angle=90, width=0.33\textwidth]{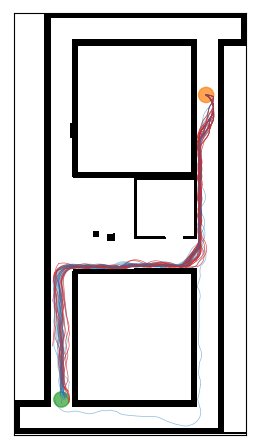}}
  \subfloat[]{\includegraphics[angle=90, width=0.33\textwidth]{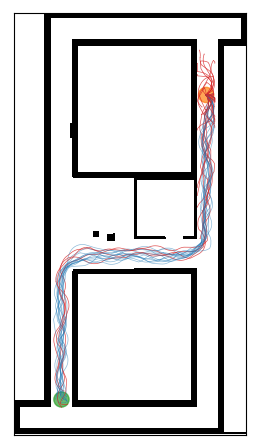}}
  \subfloat[]{\includegraphics[angle=90, width=0.33\textwidth]{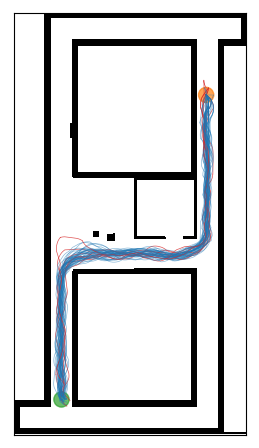}}
  \caption{(a), (b), and (c) show paths of RHC, POMCPOW, and POMCP++ with single mode initial belief, while (d), (e), and (f) are for the two mode initial belief. The orange and green patches are the start and target locations. Blue and red lines are paths of successful and failure trials. More details of the setup for the two different scenarios can be found in \figurename~\ref{fig: stochastic motion planning examples}.}
  \label{fig: paths in levine map}
\end{figure*}

In this section, we compare the performance of POMCP++ with RHC and POMCPOW in an application of navigating a differential drive ground robot in a hallway environment.
The dimension of the map is $153\times 277$, with $0.1m$ resolution.
To adapt to the enlarged map and narrow corridors, the motion primitives are changed to $\mathcal{A}_m = \{0, 0.5\}m/s \times \{-\frac{\pi}{6}\, 0, \frac{\pi}{6}\}rad/s \times \{1.0\}s$.
The radius of the goal region, $\epsilon_g$, is enlarged to $0.5m$.
Seven LIDAR beams are used as sensor measurements covering $\frac{3}{2}\pi$, with $z_{\max}=1.5m$.
As in the Sec.\ref{subsec: tests in simulated environments}, $\gamma$ is $0.99$.
The algorithms are compared in two scenarios with different initial belief shown in \figurename~\ref{fig: stochastic motion planning examples}.
The hyper-parameters for POMCPOW and POMCP++ remain the same as in Table~\ref{tab: hyperparameters for different methods}.

The comparison with DESPOT is omitted, since the robot fails to reach the goal for all trials with DESPOT.
Considering the enlarged environment compared to Sec.~\ref{subsec: tests in simulated environments}, experiments for DESPOT are also performed by increasing \textit{tree depth} to $100$ and \textit{per-step planning time} to $120s$.
There is still no successful trails for either of the initial belief setups.
The failure of DESPOT may be caused by using a single simulation episode to estimate the value at a belief node.
To elaborate, a naive application of DESPOT to systems with continuous measurement causes the degeneracy of the belief tree.
All nodes in the tree, other than the root, are only visited once, \textit{i.e.} when the node is constructed.
Therefore, only one simulation episode is available in estimating the value at the nodes, which may be far from adequate in providing an accurate estimate.
The inaccurate estimation at the nodes, eventually, leads to the wrong selection of optimal policy.
In the experiments of Sec.~\ref{subsec: tests in simulated environments}, simulation episodes are short.
It takes around $10$ steps to drive the robot to the goal.
Simulation episodes obtained with the same action sequence may not differ much.
Therefore, using only one simulation episode in estimating the value at the nodes may be sufficient and leads to meaningful actions.
However, the number of actions required to reach the goal in the hallway experiments is over $60$. The effect of degeneracy for DESPOT becomes more apparent.

Table~\ref{tab: comparison of pomcp++ and rhc} summarizes the performance of RHC, POMCPOW, and POMCP++.
For each algorithm and initial belief setup, we run the experiment $50$ times.
The paths of all trials are shown in \figurename~\ref{fig: paths in levine map}.
It is expected that RHC takes fewer actions on average for successful trials, whose planning is based on A*.
However, the total traveled distance is similar among the algorithms.
This implies that, under POMDP solutions, the robot takes more rotation actions utilizing the local environment to reduce state uncertainty.
\figurename~\ref{fig: paths in levine map} also shows that POMDP solutions are able to maintain the robot in the middle of a narrow corridor, which reduces the chance of collision caused by state uncertainty and motion stochasticity.
Comparing POMCPOW and POMCP++, although the performance of the success trails of the two algorithms is similar, POMCP++ is able to achieve higher success rate, especially when the initial belief contains ambiguity.
The improvement of POMCP++ is due to the two major modifications of POMCP, namely the new strategy of measurement selection, and the removal of the implicit perfect state assumption at the start of the rollout phase.

\begin{table*}[t]
  \centering
  \caption{Comparison of POMCPOW and POMCP++ with different per-step planning time constraints}
  \label{tab: anytime performance benchmark}
  \renewcommand{\arraystretch}{1.5}
  \begin{threeparttable}[t]
    \begin{tabular}{c|c|c|c|c|c|c}
      \hline\hline
      per-step time [$s$]
      & method
      & success/collision/wrong stop/timeout\tnote{1} [\%]
      & total actions\tnote{1}
      & total distance\tnote{1} [$m$]
      & distance to target\tnote{1} [$m$]
      & final entropy\tnote{1} \\
      \hline
      % 1s per step
      \multirow{2}{2em}{$1$}
      & POMCPOW & $    0.18 /\bm{0.18}/    0.62 /\bm{0.02}$ & $\bm{80.33}$ & $\bm{30.00}$ & $    0.40 $ & $    2.52 $\\
      & POMCP++ & $\bm{0.22}/    0.20 /\bm{0.52}/    0.06 $ & $    88.82 $ & $    30.95 $ & $\bm{0.28}$ & $\bm{2.39}$\\
      \hline
      % 5s per step
      \multirow{2}{2em}{$5$}
      & POMCPOW & $    0.64 /    0.14 /    0.20 /\bm{0.02}$ & $\bm{75.56}$ & $    30.10 $ & $    0.38$  & $\bm{2.46}$\\
      & POMCP++ & $\bm{0.74}/\bm{0.00}/\bm{0.04}/    0.22 $ & $    78.60 $ & $\bm{29.70}$ & $\bm{0.28}$ & $    2.48 $\\
      \hline
      % 10s per step
      \multirow{2}{2em}{$10$}
      & POMCPOW & $    0.68 /    0.14 /    0.10 /\bm{0.08}$ & $    75.24 $ & $    29.91 $ & $    0.39 $ & $\bm{2.54}$\\
      & POMCP++ & $\bm{0.76}/\bm{0.00}/\bm{0.04}/    0.20 $ & $\bm{74.39}$ & $\bm{29.31}$ & $\bm{0.22}$ & $    2.60 $\\
      \hline
      \multirow{2}{2em}{$15$}
      & POMCPOW & $    0.68 /    0.12 /    0.14 /\bm{0.06}$ & $    74.35 $ & $    29.70 $ & $    0.37 $ & $\bm{2.54}$\\
      & POMCP++ & $\bm{0.90}/\bm{0.02}/\bm{0.00}/    0.08 $ & $\bm{73.58}$ & $\bm{29.41}$ & $\bm{0.28}$ & $    2.67 $\\
      % 15s per step
      \hline\hline
    \end{tabular}
    \begin{tablenotes}
      \item [1] See Table~\ref{tab: comparison of pomcp++ and rhc} for the notes on the metrics.
    \end{tablenotes}
  \end{threeparttable}
\end{table*}

%\begin{figure*}
%  \centering
%  \subfloat[POMCPOW (1s)]{\includegraphics[angle=90, width=0.5\columnwidth]{./figures/pomcpow_one_sec_paths}}
%  \subfloat[POMCPOW (5s)]{\includegraphics[angle=90, width=0.5\columnwidth]{./figures/pomcpow_five_sec_paths}}
%  \subfloat[POMCPOW (10s)]{\includegraphics[angle=90, width=0.5\columnwidth]{./figures/pomcpow_ten_sec_paths}}
%  \subfloat[POMCPOW (15s)]{\includegraphics[angle=90, width=0.5\columnwidth]{./figures/pomcpow_fifteen_sec_paths}} \\
%  \subfloat[POMCP++ (1s)]{\includegraphics[angle=90, width=0.5\columnwidth]{./figures/pomcp_one_sec_paths}}
%  \subfloat[POMCP++ (5s)]{\includegraphics[angle=90, width=0.5\columnwidth]{./figures/pomcp_five_sec_paths}}
%  \subfloat[POMCP++ (10s)]{\includegraphics[angle=90, width=0.5\columnwidth]{./figures/pomcp_ten_sec_paths}}
%  \subfloat[POMCP++ (15s)]{\includegraphics[angle=90, width=0.5\columnwidth]{./figures/pomcp_fifteen_sec_paths}}
%  \caption{Paths executed by POMCPOW and POMCP++ under different one-step time budgets with the single-model belief set-up. The orange and green patches are the start and target locations. Blue and red lines are paths of successful and failure trials. More details of the setup for the scenario can be found in Fig.1 in the draft.}
%  \label{fig: anytime performance benchmark}
%\end{figure*}

We also compare the anytime performance of POMCPOW and POMCP++ in the hallway environment with single mode initial belief setup.
Table~\ref{tab: anytime performance benchmark} summarizes the performance of POMCPOW and POMCP++ under different per-step time constraints.
The simulations are conducted on a desktop computer using Intel Core i9-9920X CPU with $12$ cores running at $3.5$GHz.
The maximum per-step planning time is set to $15s$, where both POMCPOW and POMCP++ are able to finish $3000$ episodes (the default $N$ in Table~\ref{tab: hyperparameters for different methods}).
It can be observed in Table~\ref{tab: anytime performance benchmark} that POMCP++ is able to constantly outperform POMCPOW in terms of success rate.
Table~\ref{tab: anytime performance benchmark} also indicates that albeit the anytime feature, sufficient per-step planning time is required for both methods in order to achieve reasonable performance.

\subsection{Hardware Demonstration}
\label{subsec: hardware demonstration}
\begin{figure}
  \centering
  \subfloat[]{\includegraphics[width=0.23\textwidth]{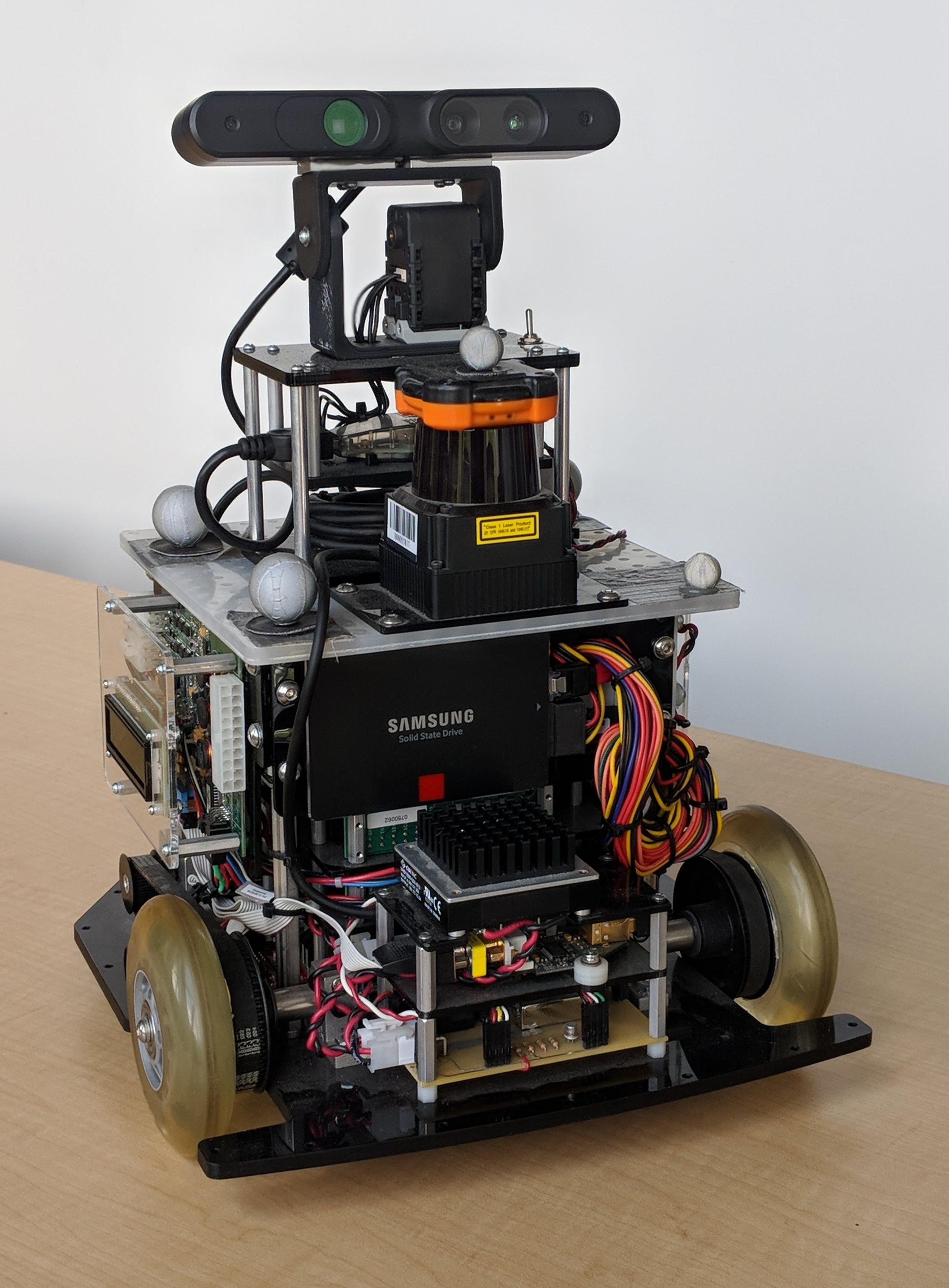}}\hfil
  \subfloat[]{\includegraphics[width=0.2075\textwidth]{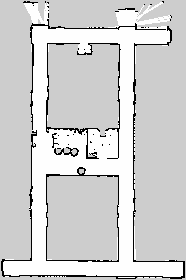}}
  \caption{(a) shows the differential drive ground robot used in the hardware experiment. Seven beams, spanning $\frac{3}{2}\pi$ with $z_{\max}=1.5m$, are used as sensor measurements. (b) is the map of the hallway environment, built with gmapping\cite{grisetti2007improved}. The dimension of the occupancy grid map, $280\times 186$ with $0.1m$ resolution, is slightly different from the one used in the simulation.}
  \label{fig: the scarab robot and the levine map}
\end{figure}
\begin{figure*}
  \centering

  \subfloat{\includegraphics[width=0.165\textwidth]{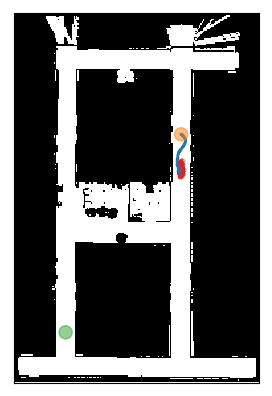}}
  \subfloat{\includegraphics[width=0.165\textwidth]{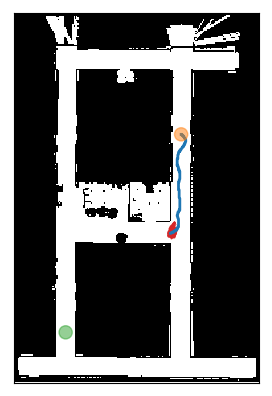}}
  \subfloat{\includegraphics[width=0.165\textwidth]{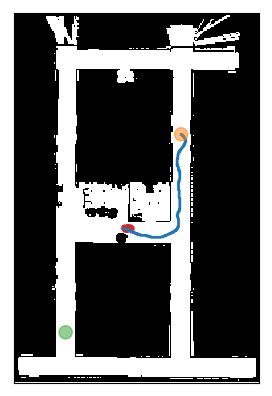}}
  \subfloat{\includegraphics[width=0.165\textwidth]{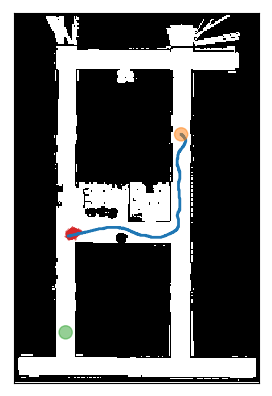}}
  \subfloat{\includegraphics[width=0.165\textwidth]{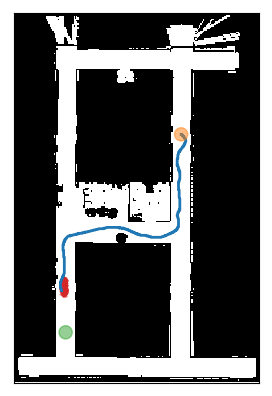}}
  \subfloat{\includegraphics[width=0.165\textwidth]{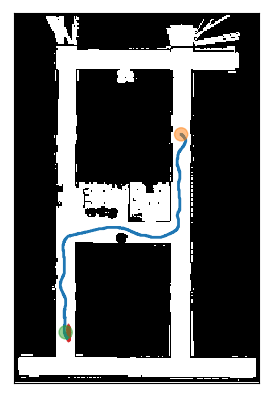}}

  \subfloat{\includegraphics[width=0.165\textwidth]{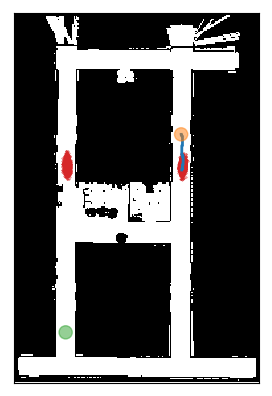}}
  \subfloat{\includegraphics[width=0.165\textwidth]{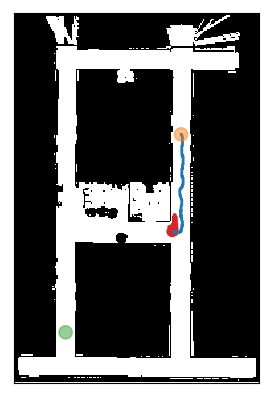}}
  \subfloat{\includegraphics[width=0.165\textwidth]{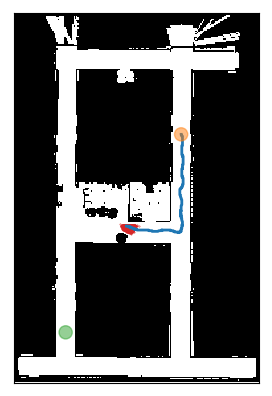}}
  \subfloat{\includegraphics[width=0.165\textwidth]{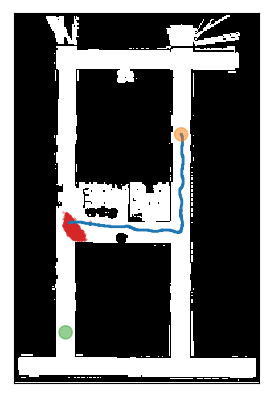}}
  \subfloat{\includegraphics[width=0.165\textwidth]{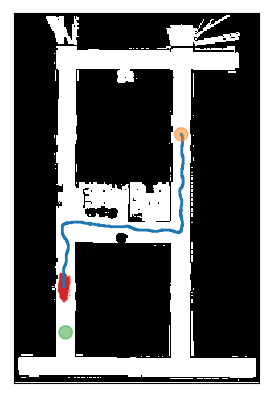}}
  \subfloat{\includegraphics[width=0.165\textwidth]{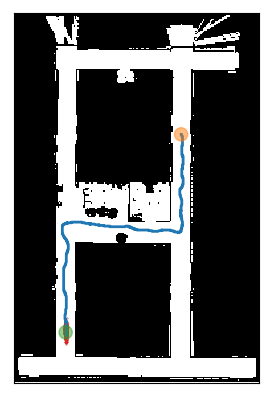}}

  \caption{Change of belief (red particles) of the robot position in the hardware experiments using POMCP++. The first row are the belief snapshots for the single mode initial belief setup, while the second row are the ones for the two mode initial belief setup. The orange and green patches are the start and target locations. Blue lines are paths estimated with an onboard LIDAR odometry.}
  \label{fig: belief change of hardware experiment in levine}
\end{figure*}
To further demonstrate the application of POMCP++, the same experiments in the simulations are carried out with hardware.
\figurename~\ref{fig: the scarab robot and the levine map} shows the setup of the ground robot and the map of the hallway environment similar to the one used in the simulations.
\figurename~\ref{fig: belief change of hardware experiment in levine} shows paths and the change of belief with the two different initial belief configurations.
It should be noted that the hardware experiment serves as a demonstration that POMCP++ can be applied to actual hardware robots and realistic environments in addition to ideal models in the simulations.
In terms of efficiency, POMCP++ may not be sufficient for real-time planning purposes, especially with onboard computers.
Most of the planning time for each step is spent on constructing rollout policies and forward simulating the system.
Using the onboard processor i7-6600U, $2$ cores running at $2.6$GHz, the planning time for an initial step can take as long as $2min$, when the robot is far from the goal.
Improving the efficiency of the algorithm is a promising direction of future research.

\section{Conclusion}
\label{sec: conclusion}
In this work, we formulate a stochastic goal navigation problem as a POMDP, and propose a new algorithm, POMCP++, a MCTS method based on POMCP for partially observed problems. The algorithm features two key improvements over prior methods, including a novel measurement selection strategy, and the usage of importance sampling to avoid over-optimism in the value estimation of the rollout policy. Compared with the LQG approaches, the proposed algorithm inherits the advantages from general POMDP solvers, such as dealing with multi-model belief, and requiring only generative models for the system. Meanwhile, with the introduced improvements, POMCP++ can be applied on systems with continuous measurements, which is often a limitation of general POMDP solvers. More importantly, the proposed POMCP++ is proved to provide unbiased estimation of the policy value with infinite number of samples and simulation episodes, thus is a valid MCTS algorithm.

We demonstrate the effectiveness of the algorithm with comparisons to both existing general purpose online POMDP solvers, and a practical robotic approach based on RHC. The results indicate that POMCP++ outperforms other methods on various metrics. In terms of the metric generally used to quantify POMDP solvers, POMCP++ achieves higher total reward. In terms of reliability in completing the navigation task, POMCP++ achieves higher success rate.

Various future research directions are promising in extending the current work. Problem-wise, it should be of broader interest to release the assumption of a known map and investigate the problem of navigation in unknown or uncertain environments. Solution-wise, improving the efficiency of POMCP++ is critical for real-time applications on computationally constrained platforms.

%\addtolength{\textheight}{-4.5cm}

%\section*{APPENDIX}
\appendices
\section{}
\label{sec: new measurement is sampled i.o.}
\begin{figure}[t]
  \centering
  \includegraphics[width=\columnwidth]{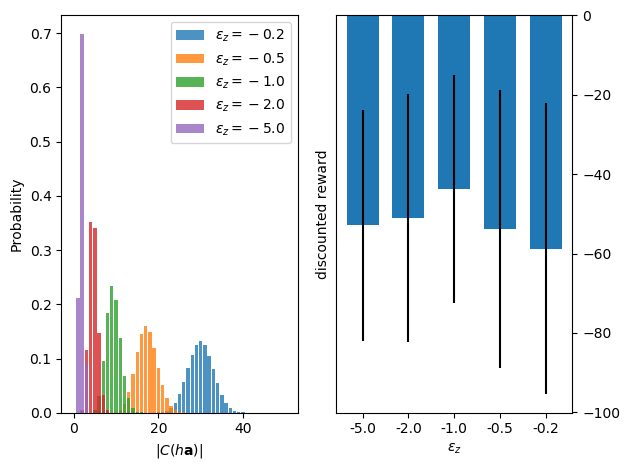}
  \caption{Left shows the distribution of $|C(h\bm{a})|$ under different $\epsilon_z$, with $N(h\bm{a})=50$. Right shows average discounted reward for different  $\epsilon_z$ on random maps of size $20\times 20$ and obstacle density $0.1$. For each $\epsilon_z$, the data is averaged over $100$ independent trials.}
  \label{fig: effect of epsilon_z}
\end{figure}

We attempt to answer two questions in this section.
First, how often a node in the tree is visited?
Second, how many child nodes a belief action node will spawn using the measurement selection strategy in Alg.~\ref{alg: pomcp++}?

\begin{lemma}
  \label{lemma: i.o. visit of belief action nodes}
  A belief action node $h\bm{a}$ will be visited \textit{i.o.} given that its parent belief node $h$ is visited \textit{i.o.} and $0<\epsilon_a<1$.
\end{lemma}
\begin{IEEEproof}
  Define $A_n$ as the event that $h\bm{a}$ is visited at the $n$\textsuperscript{th} visit of $h$.
  $A_n$'s are independent events following the definition of the action selection function in Alg.~\ref{alg: pomcp++}.
  Note $P(A_n)$ is lower bounded by $1/\epsilon_a$ with $0<\epsilon_a<1$.
  Therefore,
  \begin{equation*}
    \sum_{n=1}^{\infty} P(A_n) \geq \lim_{n\rightarrow\infty} \frac{n}{\epsilon_a} = \infty.
  \end{equation*}
  Define $A$ as the event that infinitely many $A_n$'s occur, \textit{i.e.} $A=\limsup_{n\rightarrow\infty}A_n$.
  We have $P(A)=1$ as a result of Borel-Cantelli lemmas~\cite{grimmett2020probability}.
\end{IEEEproof}

\begin{lemma}
  \label{lemma: i.o. visit of belief nodes}
  A belief node $h\bm{z}$ will be visited \textit{i.o.} given that its parent belief action node $h$ is visited \textit{i.o.} and $\epsilon_z<0$.
\end{lemma}
\begin{IEEEproof}
  Assume $h\bm{z}$ is the $m$\textsuperscript{th} spawned child node of $h$.
  Define $Z_n$ as the independent events that $h\bm{a}$ is visited at the $n$\textsuperscript{th} visit of $h$ after $h\bm{z}$ is spawned.
  $P(Z_n)$ with $n=1, 2, \dots$ is a nonincreasing sequence since the number of child nodes of $h$ is nondecreasing.
  Note $P(Z_n)$ is lower bounded by $(1-(m+n)^{\epsilon_z})/(m+n-1)$, which assumes new child node is added at every visit of $h$.
  Therefore,
  \begin{equation*}
    \begin{gathered}
      \sum_{n=1}^{\infty} P(Z_n)
      \geq \sum_{n=1}^{\infty} \frac{1-(m+n)^{\epsilon_z}}{m+n-1} \\
      \begin{aligned}
        &= \sum_{n=1}^{\infty} \frac{1}{m+n-1} - \sum_{n=1}^{\infty} \frac{1}{(m+n-1)(m+n)^{-\epsilon_z}} \\
        &= \sum_{k=m}^{\infty} \frac{1}{k} - \sum_{k=m}^{\infty} \frac{1}{k(k+1)^{-\epsilon_z}}
        \geq \sum_{k=m}^{\infty} \frac{1}{k} - \sum_{k=m}^{\infty} \frac{1}{k^{1-\epsilon_z}}.
      \end{aligned}
    \end{gathered}
  \end{equation*}
  Given $\epsilon_z<0$, $\sum_{n=1}^{\infty} P(Z_n) = \infty$ following the property of \textit{p}-series.
  We have $P(\limsup_{n\rightarrow\infty}Z_n)=1$ as a result of Borel-Cantelli lemmas.
\end{IEEEproof}

\begin{theorem}
  \label{theorem: i.o. visit of all nodes}
  If the root node is visited \textit{i.o.}, \textit{i.e.} $N\rightarrow\infty$, $0<\epsilon_a<1$, and $\epsilon_z<0$, all nodes will be visited \textit{i.o.}.
\end{theorem}
\begin{IEEEproof}
  Based on Lemma~\ref{lemma: i.o. visit of belief action nodes} and~\ref{lemma: i.o. visit of belief nodes}, the \textit{i.o.} visit of a node in the tree depends on the \textit{i.o.} visit of its parent nodes.
  Therefore, all nodes are visited \textit{i.o.} if the root node is visited \textit{i.o.}.
\end{IEEEproof}

\begin{theorem}
  \label{theorem: i.o. spawn of new belief nodes}
  A belief action node $h$ spawns child nodes \textit{i.o.} if $-1\leq \epsilon_z < 0$.
\end{theorem}
\begin{IEEEproof}
  Define $S_n$ as the independent events that a new child belief node is spawned by $h$ at its $n$\textsuperscript{th} visit.
  The sequence $P(S_n)$, $n=1, 2,\dots$ is nonincreasing, with each $P(S_n)$ lower bounded by $(n+1)^{\epsilon_z}$.
  Therefore,
  \begin{equation*}
    \sum_{n=1}^{\infty} P(S_n) \geq \sum_{n=1}^{\infty} (n+1)^{\epsilon_z}
  \end{equation*}
  A sufficient condition for $\sum_{n=1}^{\infty} P(S_n) = \infty$ is $-1 \leq \epsilon_z < 0$ ($\epsilon_z<0$ is enforced to ensure $0<(n+1)^{\epsilon_z}<1$), under which $P(\limsup_{n\rightarrow\infty}S_n)=1$, \textit{i.e.} $h$ spawns new child nodes \textit{i.o.} .
\end{IEEEproof}

Intuitively speaking, $\epsilon_z$ controls the trade-off between exploration and exploitation in measurements just as the role of $\epsilon_a$ in action selection.
\figurename~\ref{fig: effect of epsilon_z} shows distribution of $|C(h\bm{a})|$ and the discounted reward with different $\epsilon_z$.
Although the results of Theorem~\ref{theorem: i.o. spawn of new belief nodes} may not apply, cases with $\epsilon_z<-1$ are also included for comparison.

\section{}
\label{sec: convergence to optimality}
In this section, we would like to show that POMCP++ is a valid MCTS algorithm, which is, at its base, an asynchronous value iteration algorithm interleaving policy evaluation and policy improvement.
Since POMCP++ already share the algorithm structure with MCTS, the validity of POMCP++ depends on whether the value computed from the simulation episodes serves as unbiased estimates of value at the corresponding belief-action nodes.
It seems straightforward, at first glance, that estimation from simulation episodes is unbiased.
However, one should be careful in drawing the conclusion since the measurements in simulation episodes are not always obtained by forward simulating the measurement model, but are enforced to be repeated most of the time.
To fill in this gap, we start by considering the value at the belief nodes in the tree, which are not being tracked explicitly in Alg.~\ref{alg: pomcp++}.
In the following, we assume $0<\epsilon_a<1$ and $-1\leq\epsilon_z<0$ to ensure all nodes are visited \textit{i.o.}, and new child nodes are spawned \textit{i.o.} by belief action nodes as proved in Theorem~\ref{theorem: i.o. visit of all nodes} and~\ref{theorem: i.o. spawn of new belief nodes}.
\begin{lemma}
  \label{lemma: unbiased estimate at belief nodes}
  The discounted summation of rewards of the simulation episode is an unbiased estimate of the value at the traversed belief nodes as $N\rightarrow\infty$ and $K\rightarrow\infty$.
\end{lemma}
\begin{IEEEproof}
  Represent a simulation episode from belief node $h_t\in\mathcal{T}$ as $S_t \coloneqq \bm{x}_t, \bm{a}_t, \bm{z}_t, \bm{x}_{t+1}, \bm{a}_{t+1}, \bm{z}_{t+1},\dots$, with $\bm{x}_t$ sampled from $\bm{b}_t$.
  For the sake of notation sanity, define $X_t\coloneqq\bm{x}_{t:\infty}$, $A_t\coloneqq\bm{a}_{t:\infty}$, $Z_t\coloneqq\bm{z}_{t:\infty}$, i.e. $S_t=X_t, A_t, Z_t$.
  Then the value of the node $h_t$ under the policy $\pi$ is,
  \begin{equation*}
    V_{\pi}(h_t) = \int_{S_t} R(S_t) p_{\pi}(S_t) dS_t.
  \end{equation*}
  Recall $R(S_t)$ is the discounted summation of reward of a simulation episode, $p_{\pi}(S_t)$ is the \textit{p.d.f.} of $S_t$.
  Here and also in the following context, the subscript $\pi$ in $p_{\pi}(\cdot)$ indicates the \textit{p.d.f.} is related to the policy $\pi$.
  With some application of conditional probability and Bayes' rule, one has,
  \begin{equation}
    \label{eq: value of a node}
    \begin{aligned}
      V_{\pi}(h_t) &= \int_{S_t} R(S_t) p_{\pi}(S_t) dS_t \\
      &= \int_{S_t} R(S_t) p\prl{X_t | A_t, Z_t} p_{\pi}\prl{A_t, Z_t} dS_t \\
      &= \int_{S_t} p_{\pi}\prl{A_t, Z_t} R(S_t) \frac{p\prl{Z_t|X_t, A_t} p\prl{X_t|A_t}}{p\prl{Z_t|A_t}} dS_t \\
      &= \int_{A_t, Z_t} p_{\pi}\prl{A_t, Z_t}V\prl{h_t|A_t, Z_t} dA_t dZ_t,
    \end{aligned}
  \end{equation}
  where,
  \begin{equation*}
    V\prl{h_t|A_t, Z_t} = \int_{X_t} R(S_t) \frac{p\prl{Z_t|X_t, A_t} p\prl{X_t|A_t}}{p\prl{Z_t|A_t}}dX_t.
  \end{equation*}
  $V\prl{h_t|A_t, Z_t}$ is the value of the node $h_t$ given the future measurement and action sequence.
  In Alg.~\ref{alg: pomcp++}, $V\prl{h_t|A_t, Z_t}$ is estimated with,
  \begin{equation}
    \label{eq: estimation of belief nodes}
    \tilde{V}\prl{h_t|A_t, Z_t} = \frac{1}{\eta} \sum_{i=0}^{K-1} R(S_t^i) p(Z_t|X_t^i, A_t),
  \end{equation}
  where $X_t^i$ is forward sampled following the given action sequence $A_t$ with the initial state $\bm{x}_t^i$ sampled from $\bm{b}_t$.
  In~\cite{geweke1989bayesian}, Geweke proves the validity of estimation with importance sampling under weak assumptions\footnote{Briefly speaking, the four assumptions in~\cite{geweke1989bayesian} require the proper definition of prior density and the likelihood function, \textit{i.i.d.} sample sequence, and an integrable \textit{r.v.} \textit{w.r.t.} posterior density.
  All of the assumptions are satisfied in our problem setup.}, \textit{i.e.},
  \begin{equation}
    \label{eq: convergence of importance sampling}
    \tilde{V}\prl{h_t|A_t, Z_t} \xrightarrow[\text{as}\ K\rightarrow\infty]{a.s.} V\prl{h_t|A_t, Z_t},
  \end{equation}
  where $\xrightarrow{a.s.}$ means \textit{almost surely} convergence.
  Therefore,
  \begin{equation}
    \label{eq: the naive unbiased estimate}
    \int_{A_t, Z_t} p_{\pi}\prl{A_t, Z_t} \tilde{V}\prl{h_t|A_t, Z_t} dA_t dZ_t \xrightarrow[\text{as}\ K\rightarrow\infty]{a.s.} V_{\pi}(h_t).
  \end{equation}
  If one samples a measurement and action sequence from $p_{\pi}\prl{A_t, Z_t}$, Eq.~\eqref{eq: the naive unbiased estimate} shows $\tilde{V}\prl{h_t|A_t, Z_t}$ is an unbiased estimate of $V_{\pi}(h_t)$.
  However, in following the tree policy, $\prl{A_t, Z_t}$ is sampled from the distribution $p_{\mathcal{T}}\prl{A_t, Z_t}$ defined implicitly by the tree $\mathcal{T}$, instead of $p_{\pi}(A_t, Z_t)$.
  In the following, we show that if new measurements are sampled \textit{i.o.} from each belief-action node in $\mathcal{T}$, $\tilde{V}\prl{h_t|A_t, Z_t}$ still serves as an unbiased estimate of $V_{\pi}(h_t)$ even though $\prl{A_t, Z_t}$ is sampled based on $p_{\mathcal{T}}\prl{A_t, Z_t}$.

  To start, consider the first action measurement pair $\prl{\bm{a}_t, \bm{z}_t}$ in $\prl{A_t, Z_t}$.
  Define $f(\cdot)$ to be an integrable function \textit{w.r.t.} the corresponding probability measure (a bounded function defined on the sample space should suffice).
  Then,
  \begin{equation}
    \label{eq: m.i. initial step}
    \begin{gathered}
      \underset{p_{\mathcal{T}}\prl{\bm{a}_t, \bm{z}_t}}{\mathbb{E}} \crl{f(\bm{a}_t, \bm{z}_t)}
      = \int_{\bm{a}_t} \prl{\frac{1}{n}\sum_{i=0}^{n-1}f(\bm{a}_t, \bm{z}_t^i)} p_{\pi}\prl{\bm{a}_t} d\bm{a}_t \\
      \xrightarrow[\text{as}\ n\rightarrow\infty]{a.s.}
      \int_{\bm{a}_t} \prl{\int_{\bm{z}_t} f(\bm{a}_t, \bm{z}_t) p_{\pi}\prl{\bm{z}_t|\bm{a}_t}d\bm{z}_t} p_{\pi}\prl{\bm{a}_t}  d\bm{a}_t \\
      = \int_{\bm{a}_t, \bm{z}_t} f(\bm{a}_t, \bm{z}_t) p_{\pi}\prl{\bm{a}_t, \bm{z}_t} d\bm{a}_t d\bm{z}_t.
    \end{gathered}
  \end{equation}
  The first equality in Eq.~\eqref{eq: m.i. initial step} is based on how actions and measurements are selected at node $h_t\in\mathcal{T}$ in Alg.~\ref{alg: pomcp++}.
  Here and in the following context, we assume that $n$ measurements are sampled for each belief-action node in the tree with $n$ sufficiently large.
  With large $n$, we can safely ignore the probability of sampling new measurements at the belief action nodes, \textit{i.e.} one of the $n$ measurements will be selected.
  The \textit{a.s.} convergence in Eq.~\eqref{eq: m.i. initial step} follows the strong law of large numbers.

  Mathematical induction lends itself to prove the rest of the action, measurement sequence.
  Suppose,
  \begin{equation}
    \label{eq: m.i. kth step}
    \begin{aligned}
      &\underset{p_{\mathcal{T}} \prl{\bm{a}_{t:T-1}, \bm{z}_{t:T-1}}}{\mathbb{E}}
      \crl{f\prl{\bm{a}_{t:T-1}, \bm{z}_{t:T-1}}} \\
      \xrightarrow{a.s.}
      &\underset{p_{\pi} \prl{\bm{a}_{t:T-1}, \bm{z}_{t:T-1}}}{\mathbb{E}}
      \crl{f\prl{\bm{a}_{t:T-1}, \bm{z}_{t:T-1}}}.
    \end{aligned}
  \end{equation}
  Note,
  \begin{equation}
    \label{eq: m.i. k+1th step (1)}
    \begin{gathered}
      \underset{p_{\mathcal{T}} \prl{\bm{a}_{t:T}, \bm{z}_{t:T}}}{\mathbb{E}}
      \crl{f\prl{\bm{a}_{t:T}, \bm{z}_{t:T}}} \\
      =\underset{p_{\mathcal{T}} \prl{\bm{a}_{t:T-1}, \bm{z}_{t:T-1}}}{\mathbb{E}}
      \crl{\underset{p_{\mathcal{T}} \prl{\bm{a}_T, \bm{z}_T}}{\mathbb{E}}
           \crl{f\prl{\bm{a}_T, \bm{z}_T, \bm{a}_{t:T-1}, \bm{z}_{t:T-1}}}} \\
      \xrightarrow{a.s.}
      \underset{p_{\pi} \prl{\bm{a}_{t:T-1}, \bm{z}_{t:T-1}}}{\mathbb{E}}
      \crl{\underset{p_{\mathcal{T}} \prl{\bm{a}_T, \bm{z}_T}}{\mathbb{E}}
           \crl{f\prl{\bm{a}_T, \bm{z}_T, \bm{a}_{t:T-1}, \bm{z}_{t:T-1}}}}.
    \end{gathered}
  \end{equation}
  The \textit{a.s.} convergence in Eq.~\eqref{eq: m.i. k+1th step (1)} follows the assumption in Eq.~\eqref{eq: m.i. kth step}.
  It is worth mentioning that both $\underset{p_{\mathcal{T}} \prl{\bm{a}_{t:T}, \bm{z}_{t:T}}}{\mathbb{E}}\crl{f\prl{\bm{a}_{t:T}, \bm{z}_{t:T}}}$ and $\underset{p_{\pi} \prl{\bm{a}_{t:T}, \bm{z}_{t:T}}}{\mathbb{E}}\crl{f\prl{\bm{a}_{t:T}, \bm{z}_{t:T}}}$ are also bounded functions on $\prl{\bm{a}_{t:T-1}, \bm{z}_{t:T-1}}$.
  Therefore, in order to show,
  \begin{equation}
    \label{eq: m.i. k+1th step (2)}
    \begin{aligned}
      &\underset{p_{\mathcal{T}} \prl{\bm{a}_{t:T}, \bm{z}_{t:T}}}{\mathbb{E}}
      \crl{f\prl{\bm{a}_{t:T}, \bm{z}_{t:T}}} \\
      \xrightarrow{a.s.}
      &\underset{p_{\pi} \prl{\bm{a}_{t:T}, \bm{z}_{t:T}}}{\mathbb{E}}
      \crl{f\prl{\bm{a}_{t:T}, \bm{z}_{t:T}}},
    \end{aligned}
  \end{equation}
  Eq.~\eqref{eq: m.i. k+1th step (1)} suggests it is suffice to have,
  \begin{equation}
    \label{eq: m.i. k+1th step (3)}
    \begin{aligned}
      &\underset{p_{\mathcal{T}} \prl{\bm{a}_T, \bm{z}_T}}{\mathbb{E}}
      \crl{f\prl{\bm{a}_T, \bm{z}_T, \bm{a}_{t:T-1}, \bm{z}_{t:T-1}}} \\
      \xrightarrow{a.s.}
      &\underset{p_{\pi} \prl{\bm{a}_T, \bm{z}_T}}{\mathbb{E}}
      \crl{f\prl{\bm{a}_T, \bm{z}_T, \bm{a}_{t:T-1}, \bm{z}_{t:T-1}}}.
    \end{aligned}
  \end{equation}
  The same argument as Eq.~\eqref{eq: m.i. initial step} can be applied again to show the validity of Eq.~\eqref{eq: m.i. k+1th step (3)}.
  Recall that the belief action nodes in $\mathcal{T}$ are visited \textit{i.o.} and spawn new measurements \textit{i.o.} as $N\rightarrow\infty$, proved in Theorem~\ref{theorem: i.o. spawn of new belief nodes}.
  Therefore, $\tilde{V}(h_t|A_t, Z_t)$, a bounded function, is an unbiased estimate of $V_{\pi}(h_t)$ with $(A_t, Z_t)$ sampled from $p_{\mathcal{T}}(A_t, Z_t)$, \textit{i.e.},
  \begin{equation*}
    \begin{aligned}
      \underset{p_{\mathcal{T}}\prl{A_t, Z_t}}{\mathbb{E}}
      &\crl{\tilde{V}(h_t|A_t, Z_t)} &\\
      \xrightarrow[\text{as}\ N\rightarrow\infty]{a.s.}
      \underset{p_{\pi}\prl{A_t, Z_t}}{\mathbb{E}}
      &\crl{\tilde{V}(h_t|A_t, Z_t)},
      \quad &\text{Eq.~\eqref{eq: m.i. k+1th step (2)}},\\
      \xrightarrow[\text{as}\ K\rightarrow\infty]{a.s.}
      \underset{p_{\pi}\prl{A_t, Z_t}}{\mathbb{E}}
      &\crl{V(h_t|A_t, Z_t)},
      \quad &\text{Eq.~\eqref{eq: convergence of importance sampling}}, \\
      =
      V_{\pi}
      &(h_t),
      \quad &\text{Eq.~\eqref{eq: value of a node}},
    \end{aligned}
  \end{equation*}
  which completes the proof.
\end{IEEEproof}

Next we extend the conclusion to the belief action nodes in $\mathcal{T}$ as well.
\begin{theorem}
  \label{theorem: unbiased estimate at all nodes}
  The discounted summation of rewards of the simulation episode is an unbiased estimate of the value at all traversed nodes as $N\rightarrow\infty$ and $K\rightarrow\infty$.
\end{theorem}
\begin{IEEEproof}
  Given the results of Lemma~\ref{lemma: unbiased estimate at belief nodes}, it is left to show that the summation of discounted rewards is also an unbiased estimation at the traversed belief action nodes.
  Recall Eq.~\eqref{eq: estimation of belief action nodes},
  \begin{equation}
    \label{eq: estimation of belief action nodes, again}
    \tilde{V}\prl{h_t\bm{a}_t|A_{t+1}, Z_t} =
    \frac{1}{\eta} \sum_{i=0}^{K-1} p\prl{Z_t | X_t^i, A_t} R(S_t).
  \end{equation}
  Applying the theory of importance sampling again,
  \begin{equation*}
    \begin{gathered}
      \tilde{V}\prl{h_t\bm{a}_t|A_{t+1}, Z_t}
      \xrightarrow{a.s.}
      \int_{X_t} p\prl{X_t|A_t, Z_t} R(S_t) dX_t \\
       = \int_{X_t} p\prl{X_t|A_t, Z_t} \prl{r(\bm{x}_t, \bm{a}_t, \bm{x}_{t+1}) + \gamma R(S_{t+1})} dX_t,
    \end{gathered}
  \end{equation*}
  shows the summation of stage reward and reward-to-go.
  The stage reward can be simplied to,
  \begin{equation*}
    \begin{aligned}
      &\int_{X_t} p\prl{X_t|A_t, Z_t} r\prl{\bm{x}_t, \bm{a}_t, \bm{x}_{t+1}} dX_t \\
      =& \int_{\bm{x}_t, \bm{x}_{t+1}}
        r\prl{\bm{x}_t, \bm{a}_t, \bm{x}_{t+1}} p\prl{\bm{x}_t, \bm{x}_{t+1}|A_t, Z_t} d\bm{x}_t d\bm{x}_{t+1} \\
      =& r(\bm{b}_t, \bm{a}_t, \bm{b}_{t+1}).
    \end{aligned}
  \end{equation*}
  With some abuse of notation, $r(\cdot)$ denotes both the stage reward for $\bm{x}_t$, $\bm{x}_{t+1}$, and its expectation \textit{w.r.t.} $p\prl{\bm{x}_t, \bm{x}_{t+1}|A_t, Z_t}$.
  Meanwhile, the reward-to-go is,
  \begin{equation*}
    \int_{X_t} p\prl{X_t|A_t, Z_t} \cdot \gamma R(S_{t+1}) dX_t
    = \gamma \tilde{V}(h_t\bm{a}_t\bm{z}_t|A_{t+1}, Z_{t+1}).
  \end{equation*}
  Therefore, Eq.~\eqref{eq: estimation of belief action nodes, again} can be rewritten as,
  \begin{equation*}
    \tilde{V}\prl{h_t\bm{a}_t|A_{t+1}, Z_t} =
    r(\bm{b}_t, \bm{a}_t, \bm{b}_{t+1}) +
    \gamma \tilde{V}(h_t\bm{a}_t\bm{z}_t|A_{t+1}, Z_{t+1}).
  \end{equation*}
  One should be reminded that $\prl{A_{t+1}, Z_t}$ is sampled based on $p_{\mathcal{T}}\prl{A_{t+1}, Z_t}$.
  Then, the expectation of $\tilde{V}\prl{h_t\bm{a}_t|A_{t+1}, Z_t}$ is,
  \begin{equation*}
    \begin{gathered}
      \underset{p_{\mathcal{T}}\prl{A_{t+1}, Z_t}}{\mathbb{E}}
      \crl{\tilde{V}\prl{h_t\bm{a}_t|A_{t+1}, Z_t}} =\\
      r(\bm{b}_t, \bm{a}_t, \bm{b}_{t+1}) +
      \gamma \sum_{i=0}^{n-1}\underset{p_{\mathcal{T}}\prl{A_{t+1}, Z_{t+1}}}{\mathbb{E}}
      \crl{\tilde{V}(h_t\bm{a}_t\bm{z}_t^i | A_{t+1}, Z_{t+1})} \\
      \xrightarrow{a.s.}
      r(\bm{b}_t, \bm{a}_t, \bm{b}_{t+1}) +
      \gamma \sum_{i=0}^{n-1} V_{\pi}\prl{h_t\bm{a}_t\bm{z}_t^i}
      \xrightarrow{a.s.}\\
      r(\bm{b}_t, \bm{a}_t, \bm{b}_{t+1}) +
      \gamma \int_{\bm{z}_t} V_{\pi}\prl{h_t\bm{a}_t\bm{z}_t^i} p\prl{\bm{z}_t|\bm{a}_t}d\bm{z}_t
      = V_{\pi}\prl{h_t\bm{a}_t}
    \end{gathered}
  \end{equation*}
  The first \textit{a.s.} convergence is based on Lemma~\ref{lemma: unbiased estimate at belief nodes}, while the second is, again, from the strong law of large numbers.
  Therefore, $\tilde{V}\prl{h_t\bm{a}_t|A_{t+1}, Z_t}$ is an unbiased estimate of $V_{\pi}\prl{h_t\bm{a}_t}$.
\end{IEEEproof}

Comparing Eq.~\eqref{eq: estimation of belief nodes} and Eq.~\eqref{eq: estimation of belief action nodes, again}, one might find it striking that $\tilde{V}\prl{h_t\bm{a}_t|A_{t+1}, Z_t}=\tilde{V}\prl{h_t|A_t, Z_t}$.
The results should not be over-interpreted leading to the conclusion that $V_{\pi}(h_t\bm{a}_t) = V_{\pi}(h_t)$.
The correct statement is $\frac{1}{\eta}\sum_i^{K-1}P(Z_t|X_t^i, A_t)R(S_t^i)$ serves as an unbiased estimate for both $V_{\pi}(h_t\bm{a}_t)$ and $V_{\pi}(h_t)$.
The discomfort should be resolved by considering that $V_{\pi}(h_t)$ is updated (implicitly) when any of its child belief-action nodes is visited, while updating $V_{\pi}(h_t\bm{a}_t)$ requires $h_t\bm{a}_t$, one child of $h_t$, to be visited.

The immediate consequence of Theorem~\ref{theorem: unbiased estimate at all nodes} is that Alg.~\ref{alg: pomcp++} is a \textit{bona fide} MCTS algorithm.
To the best of our knowledge, there is no formal proof yet that MCTS would converge to the optimal action at the root when applied to stochastic systems.
Kocsis~\cite{kocsis2006bandit} presents conclusions on the convergence of UCT.
Unfortunately, the details of the proof are not available.
On a related note, it is also commented in~\cite[Ch.5]{sutton2018reinforcement} that proving the convergence of Monte Carlo Control, a more general algorithm that MCTS bases on, could be a fundamental open question in reinforcement learning.

%\section*{ACKNOWLEDGMENT}
%We gratefully acknowledge the support of ARL grants W911NF-08-2-0004, W911NF-17-2-0181 and DARPA grants HR001151626, HR0011516850.

% Bibliography
%==================================================================%
\bibliographystyle{IEEEtran}
\bibliography{ref}
%==================================================================%

\addtolength{\textheight}{-7.5cm}
\begin{IEEEbiography}[{\includegraphics[width=1in,height=1.25in,clip,keepaspectratio]{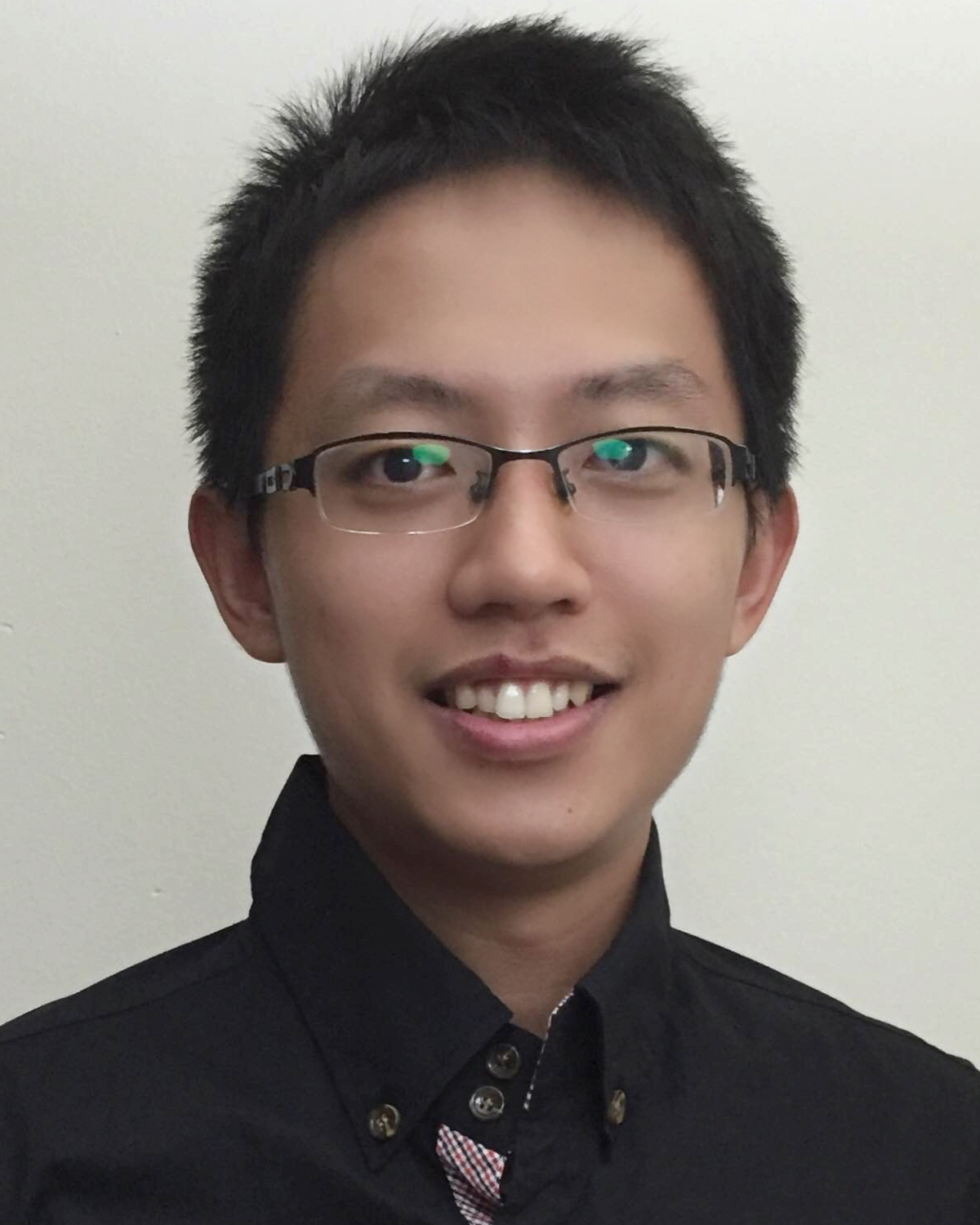}}]{Ke Sun}
  (S'20) holds a B.E. degree in electronics and information engineering from The Hong Kong Polytechnic University, Hong Kong, China in 2013 and a M.S. degree in electrical and computer engineering from Carnegie Mellon University, Pittsburg, PA, USA, in 2015.
  He is currently working towards the Ph.D. in electrical and systems engineering department at University of Pennsylvania, PA, USA.

  His research interests include state estimation and stochastic motion planning with applications to autonomous vehicles.
\end{IEEEbiography}
\begin{IEEEbiography}[{\includegraphics[width=1in,height=1.25in,keepaspectratio]{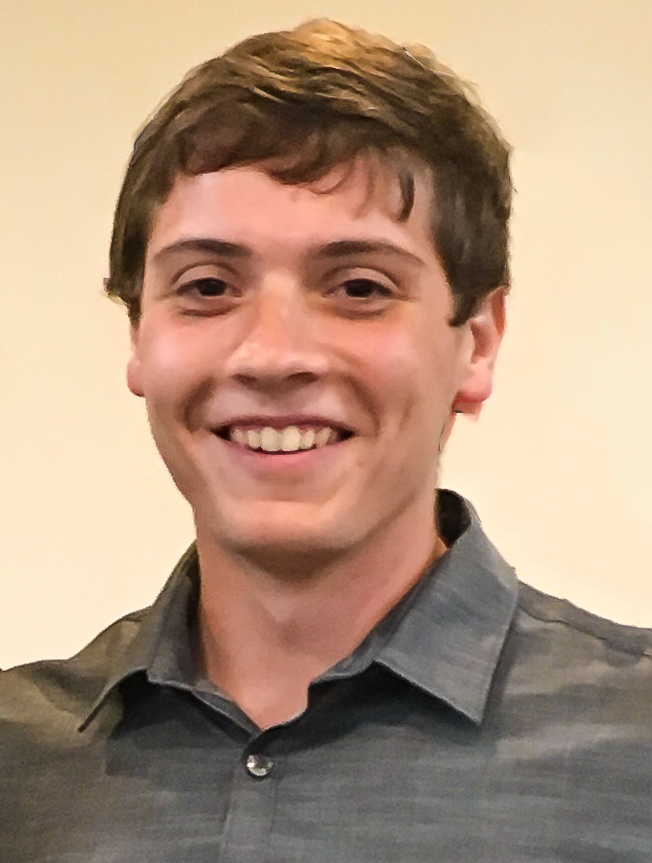}}]{Brent Schlotfeldt}
  (S'16) holds a Master of Science in robotics from the University of Pennsylvania, Philadelphia, PA (2017), and a Bachelor of Science in electrical engineering and computer science from the University of Maryland, College Park, MD, USA (2016).
  Currently, He is working towards his Ph.D. in Electrical and Systems Engineering at the University of Pennsylvania, Philadelphia, PA, USA.

  His research interests include planning, control, and state estimation for robotic systems, with applications to autonomous driving and active sensing.
\end{IEEEbiography}
\begin{IEEEbiography}[{\includegraphics[width=1in,height=1.25in,clip,keepaspectratio]{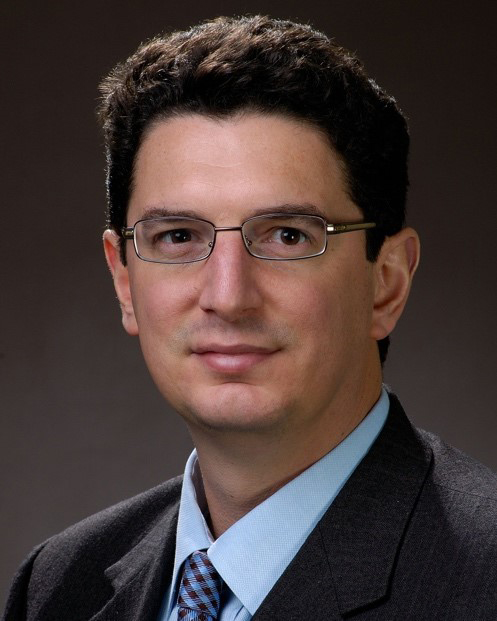}}]{George Pappas}
  (F’09) received the Ph.D. degree in electrical engineering and computer sciences from the University of California, Berkeley, CA, USA, in 1998.

  He is currently the Joseph Moore Professor and Chair of the Department of Electrical and Systems Engineering, University of Pennsylvania, Philadelphia, PA, USA.
  He also holds a secondary appointment with the Department of Computer and Information Sciences and the Department of Mechanical Engineering and Applied Mechanics.
  He is a Member of the GRASP Lab and the PRECISE Center.
  He had previously served as the Deputy Dean for Research with the School of Engineering and Applied Science.
  His research interests include control theory and, in particular, hybrid systems, embedded systems, cyber-physical systems, and hierarchical and distributed control systems, with applications to unmanned aerial vehicles, distributed robotics, green buildings, and bimolecular networks.

  Dr. Pappas has received various awards, such as the Antonio Ruberti Young Researcher Prize, the George S. Axelby Award, the Hugo Schuck Best Paper Award, the George H. Heilmeier Award, the National Science Foundation PECASE Award and numerous Best Student Papers Awards at the American Control Conference, the IEEE Conference on Decision and Control, and the ACM/IEEE International Conference on Cyber-Physical Systems.
\end{IEEEbiography}
\begin{IEEEbiography}[{\includegraphics[width=1in,height=1.25in,clip,keepaspectratio]{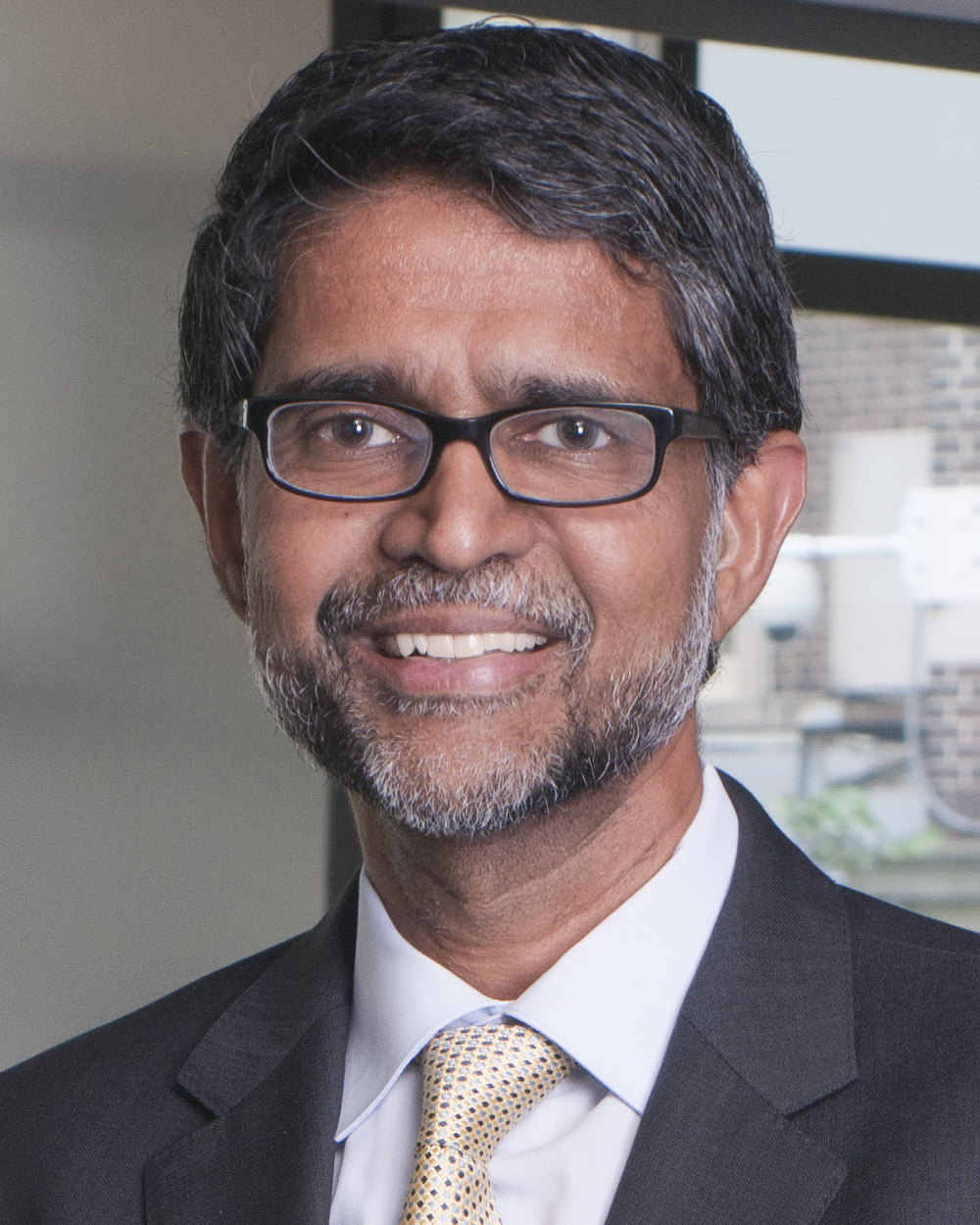}}]{Vijay Kumar}
  (F’05) received the Ph.D. degree in mechanical engineering from The Ohio State University, Columbus, OH, USA, in 1987.

  He is the Nemirovsky Family Dean of Penn Engineering (with appointments) with the Department of Mechanical Engineering and Applied Mechanics, the Department of Computer and Information Science, and the Department of Electrical and Systems Engineering, University of Pennsylvania, Philadelphia, PA, USA.
  He served as the Assistant Director of Robotics and Cyber Physical Systems with the White House Office of Science and Technology Policy from 2012 to 2014.
  He is the Founder of Exyn Technologies, a company that develops solutions for autonomous flight.

  Dr. Kumar became a Fellow of the American Society of Mechanical Engineers in 2003 and a member of the National Academy of Engineering in 2013.
  He is the recipient of the 2012 World Technology Network Award, the 2013 Popular Science Breakthrough Award, the 2014 Engelberger Robotics Award, the 2017 IEEE Robotics and Automation Society George Saridis Leadership Award in Robotics and Automation, and the 2017 ASME Robert E. Abbott Award.
  He has received best paper awards at the 2002 International Symposium on Distributed Autonomous Robotic Systems (DARS), the 2004 IEEE International Conference on Robotics and Automation (ICRA), ICRA 2011, the 2011 Robotics: Science and Systems Conference (RSS), RSS 2013, and the 2015 EAI International Conference on Bio-Inspired Information and Communications Technologies.
  He has advised doctoral students, who have received best student paper awards at ICRA 2008, RSS 2009, and DARS 2010.
\end{IEEEbiography}

\end{document}